\documentclass{article}

\usepackage[final]{neurips_2025}

\usepackage[utf8]{inputenc} 
\usepackage[T1]{fontenc}    
\usepackage{hyperref}       
\usepackage{url}            
\usepackage{booktabs}       
\usepackage{amsfonts}       
\usepackage{nicefrac}       
\usepackage{microtype}      
\usepackage{xcolor}         
\usepackage{graphicx}
\usepackage{wrapfig}
\usepackage{caption}
\usepackage[table]{xcolor}
\usepackage{pifont}
\newcommand{\cmark}{\textcolor{green!60!black}{\ding{51}}} 
\newcommand{\xmark}{\textcolor{red}{\ding{55}}}            %
\usepackage{enumitem}  
\usepackage{multirow}
\usepackage{makecell}
\usepackage{amsmath}
\usepackage{arydshln}
\definecolor{sharedcolor}{HTML}{BFBFFF}
\definecolor{GSE}{HTML}{FFFFBF}
\definecolor{LPR}{HTML}{BFFFFF}
\title{\textsc{FineRS}: Fine-grained Reasoning and Segmentation of Small Objects with Reinforcement Learning}

\author{%
  Lu Zhang$^1$\thanks{Corresponding authors}, Jiazuo Yu$^1$, Haomiao Xiong$^1$, Ping Hu$^2$, Yunzhi Zhuge$^1$\footnotemark[1], Huchuan Lu$^1$, You He$^3$ \\
  $^1$Dalian University of Technology, Dalian, China\\
  $^2$University of Electronic Science and Technology of China, Chengdu, China\\
  $^3$Tsinghua Shenzhen International Graduate School, Shenzhen, China \\
  \texttt{zhangluu@dlut.edu.cn}, yujiazuo@mail.dlut.edu.cn \\
  \url{https://iiau-zhanglu.github.io/FINERS/}\\
}

\begin{document}

\maketitle

\begin{abstract}

Multi-modal Large Language Models (MLLMs) have shown remarkable capabilities across a wide range of vision-language tasks. However, due to the restricted input resolutions, MLLMs face significant challenges in precisely understanding and localizing visual details in high-resolution images---particularly when dealing with extra-small objects embedded in cluttered contexts. 
To address this issue, we propose \textsc{FineRS}, a two-stage MLLM-based reinforcement learning framework for jointly reasoning and segmenting extremely small objects within high-resolution scenes. \textsc{FineRS} adopts a coarse-to-fine pipeline comprising Global Semantic Exploration (GSE) and Localized Perceptual Refinement (LPR). Specifically, GSE performs instruction-guided reasoning to generate a textural response and a coarse target region, while LPR refines this region to produce an accurate bounding box and segmentation mask. To couple the two stages, we introduce a locate-informed retrospective reward, where LPR's outputs are used to optimize GSE for more robust coarse region exploration.   
Additionally, we present \textsc{FineRS}-4k, a new dataset for evaluating MLLMs on attribute-level reasoning and pixel-level segmentation on subtle, small-scale targets in complex high-resolution scenes. Experimental results on \textsc{FineRS}-4k and public datasets demonstrate that our method consistently outperforms state-of-the-art MLLM-based approaches on both instruction-guided segmentation and visual reasoning tasks.

\end{abstract}
\section{Introduction}

\begin{figure}[t]
  \centering
  \includegraphics[width=1\linewidth]{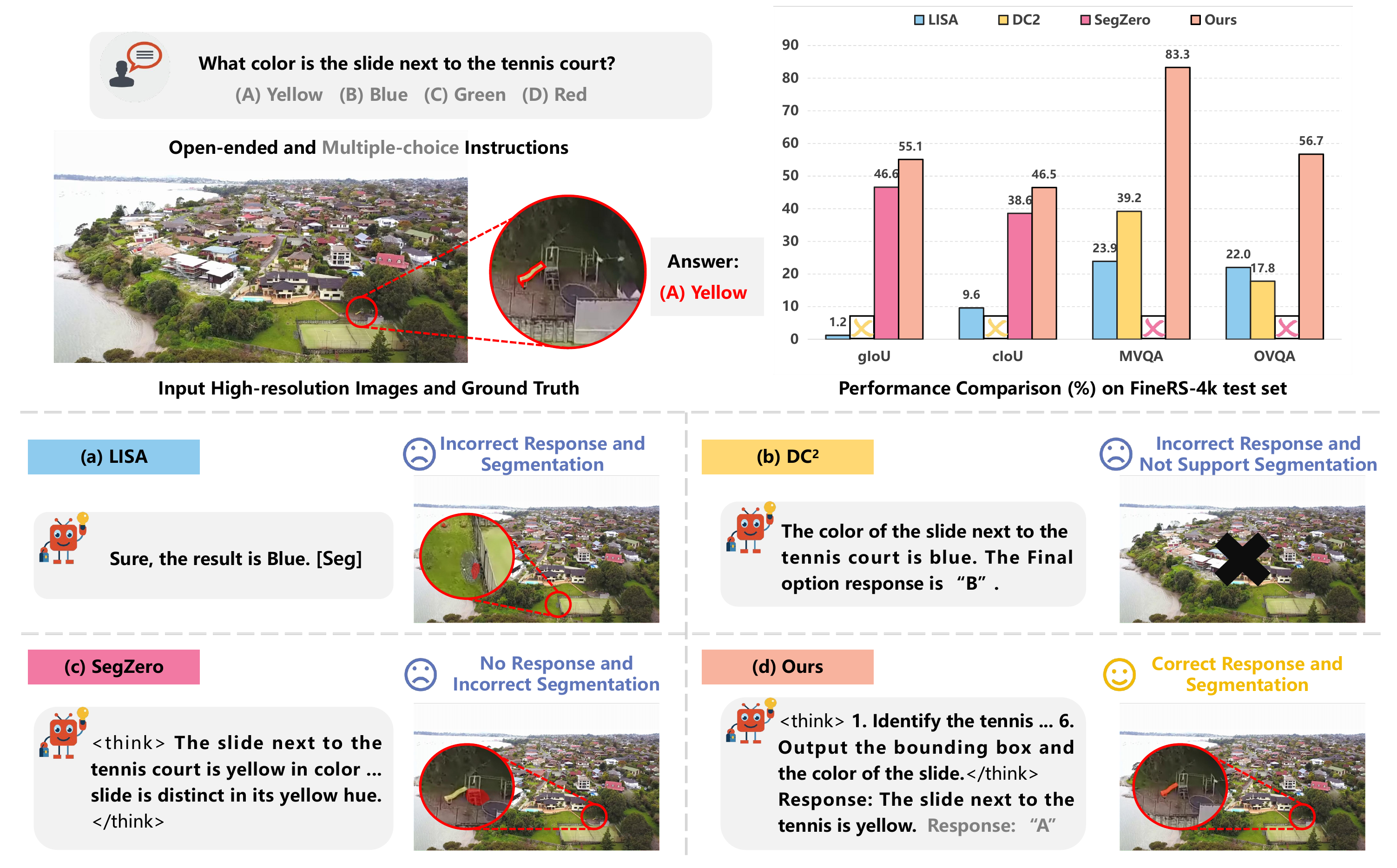}
  \caption{Given a user instruction and a high-resolution image, existing MLLMs face limitations in effectively reasoning and segmenting small objects. (a) MLLMs (e.g., LISA~\cite{lai2024lisa}) designed for standard-resolution images fail to generalize to small objects. (b) MLLMs (e.g., DC$^2$~\cite{wang2025divide}) developed for high-resolution images only support optional question answering and lack localization ability. (c) MLLMs (e.g., SegZero~\cite{liu2025segzero}) with RFT fail to produce explicit answers and accurate segmentation masks within a unified framework. (d) Our \textsc{FineRS} combines a coarse-to-fine perception pipeline with reinforcement learning, enabling unified, precise instruction-guided reasoning and segmentation of small objects in high-resolution images.}
  \label{fig:teaser}
  \vspace{-4mm}
\end{figure}

Recently, Multi-modal Large Language Models (MLLMs)~\cite{zhang2025mllms,yang2024qwen2, lai2024lisa, liu2023visual} have achieved remarkable success in a variety of vision-language tasks, such as visual question answering, referring expression comprehension, and instruction-guided segmentation. Among these tasks, one foundational challenge is instruction-guided reasoning and segmentation --- {a capacity of not only understanding what the user is asking, but also where in the image the referred object appears at the pixel level.}  
Some early attempts~\cite{lai2024lisa} integrate MLLMs~\cite{liu2023visual, touvron2023llama, chiang2023vicuna} with foundational segmentation models~\cite{kirillov2023sam,ravi2024sam2}, enabling joint language generation and object segmentation for more interactive and interpretable visual understanding. However, these methods are tailored to standard-resolution images and large, prominent objects, where spatial structures are easily accessible to the model's visual backbone. Their heavy reliance on global visual-semantic alignment becomes increasingly unreliable in scenes with dense layouts and small, low-saliency targets (see Fig.~\ref{fig:teaser} (a)).  

To address this challenge, the research community has begun to explore the capacity of MLLMs to perceive fine-grained, detailed objects within high-resolution images. To mitigate detail degradation caused by image downsampling, existing methods~\cite{zhang2025mllms, wu2024v, wang2025divide} mimic human visual perception by decomposing high-resolution images into smaller patches to achieve local vision-text alignment. However, due to the scarcity of high-resolution data, these methods typically adopt a training-free pipeline, where the absence of supervised fine-tuning limits their perception accuracy in complex scenarios (as shown in Fig.~\ref{fig:teaser} (b)). More importantly, the lack of precise localization ability restricts their scalability and applicability to downstream tasks demanding pixel-level grounding and spatial reasoning. 

Recent studies~\cite{jaech2024openai,guo2025deepseek,shao2024deepseekmath} have revealed that LLMs can generalize effectively to domain-specific tasks with only thousands of training samples. Moreover, incorporating a ``thinking'' process prior to answering can significantly enhance their reasoning ability. The core technique behind this improvement is Reinforcement Fine-Tuning (RFT)~\cite{guo2025deepseek,liu2025visualrft,shao2024deepseekmath}, which enables LLMs to be emergently optimized for downstream tasks via data-efficient fine-tuning. The success of RFT has driven the extension of vision RFT~\cite{liu2025visualrft, huang2025vision, liu2025segzero} to empower MLLMs across a variety of vision-language tasks, including image classification~\cite{liu2025visualrft}, object detection~\cite{liu2025visualrft}, and reasoning segmentation~\cite{liu2025segzero}. 
However, due to input resolution limitations, these methods~\cite{liu2025segzero} still struggle to capture fine-grained details, and cannot simultaneously generate explicit answers and segmentation masks without a unified multi-task reward mechanism (see Fig.~\ref{fig:teaser} (c)). 

To address the above issues, we propose \textsc{FineRS}, a two-stage MLLM framework for instruction-guided reasoning and segmentation of small objects in high-resolution images. \textsc{FineRS} is designed to jointly optimize semantic reasoning and spatial perception through a coarse-to-fine pipeline. Specifically, a Global Semantic Exploration (GSE) module first performs instruction-guided reasoning to produce both a textual response and a coarse target region that contains the referred objects inside. The subsequent Localized Perceptual Refinement (LPR) module then refines this region by generating an accurate bounding box and the corresponding segmentation mask.
To reduce reliance on heavy supervised fine-tuning, we integrate vision RFT into our framework and design effective rewards to simultaneously address object reasoning and segmentation. Specifically, in addition to basic region/box regularization rewards, we introduce a response reward to encourage the model to simultaneously generate both textual answers and object boxes. Furthermore, 
to effectively couple these two stages, we introduce a locate-informed retrospective reward, which uses LPR to guide and optimize the exploration behavior of GSE via reinforcement learning. 
The synergy between the coarse-to-fine framework and reinforcement learning not only enables fine-grained perception of extremely small objects but also allows for data-efficient multi-task training, resulting in a unified framework that consistently delivers high performance across both reasoning and segmentation tasks.  

To enable a comprehensive evaluation, we introduce \textsc{FineRS}-4k, a human-annotated, high-quality dataset designed to benchmark model performance on Instruction-guided Segmentation (IS), Open-ended Visual Question Answering (OVQA), and Multiple-choice Visual Question Answering (MVQA).
Compared to previous high-resolution benchmarks~\cite{wu2024v, wang2025divide}, \textsc{FineRS}-4k leverages UAV-captured imagery, providing large-scale, complex environments with extreme object size variability, scattered small-object distributions, and cluttered spatial contexts. Extensive experiments on \textsc{FineRS}-4k and other public datasets~\cite{wu2024v, wang2025divide} demonstrate that \textsc{FineRS} consistently outperforms existing MLLM-based methods in both answer accuracy and segmentation precision. 

To summarize, our contributions are as follows:
\begin{itemize}[leftmargin=*]
    \item We propose \textsc{FineRS}, a two-stage MLLM framework that jointly performs instruction-guided reasoning and segmentation for small-object understanding in high-resolution images. To the best of our knowledge, \textsc{FineRS} is the first method to unify reasoning and fine-grained segmentation under a reinforcement learning paradigm.
    \item We introduce \textsc{FineRS}-4k, the first UAV-captured high-resolution dataset designed for instruction-guided reasoning and segmentation on ultra-small objects, offering more challenging object distributions and spatial variability compared to previous datasets. 
    \item We conduct extensive experiments on \textsc{FineRS}-4k and other public datasets, demonstrating that the proposed \textsc{FineRS} consistently outperforms state-of-the-art MLLM-based approaches across instruction-guided segmentation, open-ended VQA, and multiple-choice VQA.
\end{itemize}

\section{Related Works}
\textbf{MLLM-based Reasoning and Segmentation.}
The success of MLLMs has significantly advanced object detection and segmentation for more accurate open-world understanding. Pioneering efforts such as LISA~\cite{lai2024lisa} and LISA++~\cite{yang2023lisa++} introduce a \texttt{<SEG>} token to bridge MLLMs with segmentation models~\cite{kirillov2023sam}. To evaluate performance, they propose reasoning segmentation, an extension of referring segmentation that enables simultaneous generation of text responses and segmentation masks. This design has inspired a series of follow-up works~\cite{bai2024one,ren2024pixellm}, which further explore special-token-based interfaces to integrate vision-language reasoning with segmentation. Despite their promising results, these approaches still face notable limitations. First, most MLLM-based models restrict the input resolution to avoid out-of-memory risks, leading to severe downsampling that compromises fine-grained visual details and degrades performance on high-resolution imagery. Second, they rely heavily on supervised training with extensive public datasets~\cite{mao2016generation, lai2024lisa}, which limits their adaptability and transferability to more challenging scenarios with limited training data availability.
In this work, we aim to address the challenges of small-object reasoning and segmentation within high-resolution images. We propose a two-stage MLLM framework that combines global semantic exploration with localized perceptual refinement and applies a reinforcement learning strategy to achieve data-efficient fine-tuning.

\textbf{High-resolution Image Understanding and Reasoning.}
Recent studies~\cite{wu2024v,wang2025divide,zhang2025mllms,shen2024zoomeye} have revealed that MLLMs still face significant challenges in perceiving and reasoning over high-resolution images, particularly for small and densely distributed objects.
To overcome the resolution restrictions of MLLMs, fine-tuning-based methods~\cite{liu2024improved, chen2024internvl} divide input images into uniform patches and process them in parallel with visual encoders, enabling MLLMs to handle arbitrary-resolution inputs.
Concurrently, SEAL~\cite{wu2024v} considers a more complex challenge of small object perception in high-resolution images. It introduces both an evaluation benchmark and a guided visual search mechanism that leverages LLM priors to selectively focus on important regions, effectively improving visual reasoning ability in complex and crowded high-resolution scenarios. Due to the scarcity of available training data, subsequent methods~\cite{wang2025divide,shen2024zoomeye} have developed training-free pipelines that apply hierarchical image partitioning to form stepwise reasoning processes. Similarly, attention-based visual intervention methods~\cite{zhang2025mllms} enhance the perception of small visual details by interpreting and manipulating internal attention maps of MLLMs. Despite these advancements, existing methods remain sensitive to heuristic cropping algorithms and still fail to achieve precise object localization. In this paper, we introduce a large-scale dataset and propose a data-efficient fine-tuning pipeline that employs reinforcement learning to adapt MLLMs for small object reasoning and segmentation. 
\textbf{Reinforcement Learning for MLLMs.}
Recently, Reinforcement Learning (RL)~\cite{cao2024survey, gu2024review} has become a new-emerging technique for enhancing reasoning in large language models, as demonstrated by OpenAI's o1~\cite{jaech2024openai} and DeepSeek R1-Zero~\cite{guo2025deepseek}. Among them, a critic-free algorithm, Group Relative Policy Optimization (GRPO)~\cite{shao2024deepseekmath} is designed to eliminate Supervised Fine-Tuning (SFT) by directly comparing candidate responses in groups.
Inspired by this, Visual-RFT~\cite{liu2025visualrft} proposes an RL-based fine-tuning strategy for Large Vision-Language Models (LVLMs), improving performance on classification and detection tasks using GRPO-based rewards under limited supervision.
However, it is limited to coarse-level tasks such as classification and detection, and does not support fine-grained segmentation. In contrast, Seg-Zero~\cite{liu2025segzero} leverages high-quality box-level rewards within an RL framework and feeds them into a frozen SAM2~\cite{ravi2024sam2} segmentation model, enabling pixel-level visual perception and reasoning. While it performs well on standard-resolution and regular object scenarios, challenges remain in small-object segmentation for high-resolution images. Moreover, Seg-Zero adopts a fixed reward paradigm, making it difficult to generalize to more diverse question-answering formats, such as open-ended and multiple-choice VQA. 
Inspired by the success of RFT in vision-language tasks, we apply RFT to our two-stage MLLM framework, where both localization and VQA rewards are integrated to boost the unification of object segmentation and VQA. Besides, we introduce a retrospective reward between two stages for more consistent global-to-local perception.  

\begin{table*}[t]
\scriptsize
\centering

\caption{Comparisons of different benchmarks. The annotation type includes Question (Q), Answer (A), and object mask. The supported tasks are Multiple-choice Visual Question Answering (MVQA), Open-ended Visual Question Answering (OVQA), and Instruction-guided Segmentation (IS). For small object granularity, ``Partial'' means that the dataset contains some small objects but doesn't provide an explicit indication.}
\begin{tabular}{|l
    >{\centering\arraybackslash}m{1.2cm} 
    >{\centering\arraybackslash}m{1.8cm}
    >{\centering\arraybackslash}m{1.5cm} 
    >{\centering\arraybackslash}m{2.2cm} 
    >{\centering\arraybackslash}m{2.8cm}|
}
\hline
\rowcolor{gray!35} \textbf{Dataset} &  \textbf{HR Images} & \textbf{Annotation Type} &\textbf{Sample Num}& \textbf{Supported Tasks} &\textbf{Small Object Granularity} \\ 
\hline
\hline

V$^*$~\cite{wu2024v} & \cmark & Q$\&$A &191&  MVQA & \xmark \\
HR-Bench~\cite{wang2025divide}& \cmark & Q$\&$A &200 &MVQA &  \xmark \\
refCOCOg~\cite{mao2016generation} & \xmark & Q$\&$Mask &95,010&  IS & Partial   \\
ReasonSeg~\cite{lai2024lisa}& \xmark & Q$\&$Mask &1,218&  IS& Partial   \\
\hline
\hline
\textbf{Ours (\textsc{FineRS}-4k)}&\cmark  & \textbf{Q$\&$A$\&$Mask} &\textbf{12,132}& \textbf{MVQA$\&$OVQA$\&$IS} & \textbf{S/XS/XXS}  \\
\hline
\end{tabular}
\label{tab:bench_compare}
\end{table*}

 \begin{figure}[t]
  \centering
  \includegraphics[width=1\linewidth]{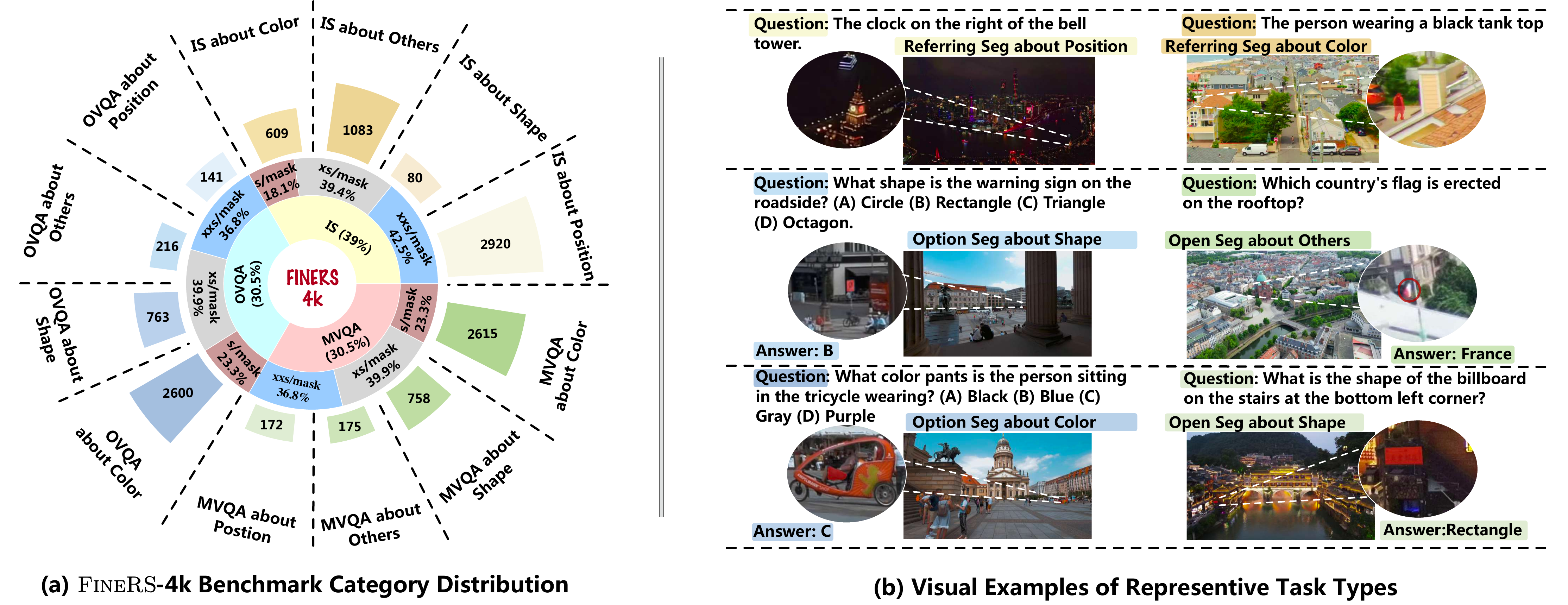}
  \caption{Overview of our benchmark. (a) The innermost ring shows three instruction types. The middle ring presents the mask size distribution within each type. The outermost ring breaks each type down into four attribute categories: color, shape, position, and others. (b) Visualization of six representative examples, each illustrating a different attribute category (color, shape, position, others) across the three instruction types in the outermost ring.}
  
  \label{fig:benchmark}
  \vspace{-4mm}
\end{figure}

\section{\textsc{FineRS}-4k Benchmark}

To comprehensively evaluate the capacities of MLLMs for instructed-guided ultra-small object reasoning and segmentation, we construct a new dataset \textsc{FineRS}-4k. Unlike previous high-resolution benchmarks~\cite{wu2024v,wang2025divide} that are primarily captured by handheld or ground-based cameras under structured conditions, \textsc{FineRS}-4k comprises images captured by Unmanned Aerial Vehicles (UAVs). This enables a much wider field-of-view and introduces complex visual challenges such as dense layouts, extreme variations in object scale, and small-object sparsity. 
%

We begin by collecting 4k-resolution drone videos ($3840\times2060$) from YouTube and our own UAV footage. Volunteers are then tasked with filtering high-quality frames and annotating small objects with a triplet annotation consisting of question, answer, and mask. 
Considering the difficulty in annotating ultra-small objects in high resolution images, annotators were primarily instructed to identify a single, unambiguous small object of interest from the image and compose questions that uniquely specify it (e.g., referencing color, shape, position, or context). For several cases with multiple similar objects, they were required to formulate disambiguating questions and to visually inspect the entire image to ensure no other object matched the same description.
The annotation was completed by 14 volunteers, organized in pairs for mutual cross-checking. In addition, a team of 4 senior reviewers conducted a final round of quality assurance to correct ambiguities and verify consistency. This multi-stage validation process was designed to maximize precision and minimize annotation bias.
This process results in 8,411 annotated small entities across 4,563 high-resolution images, yielding a total of 12,132 text-mask pairs. Specifically, we divide them into train set (8,956), validation set (749), and test set (2,427). The overall comparison of our \textsc{FineRS}-4k and other datasets is illustrated in Tab.~\ref{tab:bench_compare}.  

Fig.~\ref{fig:benchmark} illustrates a detailed analysis of the distribution of task type and object sizes. As shown in the innermost ring of Fig.~\ref{fig:benchmark} (a), \textsc{FineRS}-4k provides instructions for three sub-tasks, including \textbf{1) Instruction-guided Segmentation (IS, 39\%)} that requires generating a mask based on the instruction; \textbf{2) Multiple-choice VQA (MVQA, 30.5\%)} that involves predicting both a segmentation mask and an option based on the option-given instruction; and \textbf{3) Open-ended VQA (OVQA, 30.5\%)} that requires producing both a mask along and a free-form answer. Each entity is bound to at least one instruction type. We further classify object size into three categories based on their proportion to the entire image area: small (S, >0.055\%), extra small (XS, 0.017\%–0.055\%), and extra-extra small (XXS, < 0.017\%). The distribution of these object sizes across task types is shown in the second ring of Fig.~\ref{fig:benchmark} (a). The outermost ring illustrates the types of attribute-specific instructions, including color, shape, position, and other distributions. Combining the three task types and four attribute types yields 12 distinct instruction-task combinations, with the numbers indicating the annotated instance count for each sub-task. Fig.~\ref{fig:benchmark} (b) illustrates the visual examples of different tasks in \textsc{FineRS}-4k. More detailed analysis about object size and spatial distribution can be found in Fig.~\ref{Ap:spatial_distribution}.
\section{Methodology}


\subsection{Preliminary} 

\textbf{Task Definition.}
Given an image $I$ and a user instruction $Q$, MLLMs~\cite{liu2023visual,yang2024qwen2} are capable of jointly understanding visual and textual input to generate appropriate responses $A^{pre}$. The objective of instruction-guided segmentation~\cite{lai2024lisa,liu2025segzero} is to predict a segmentation mask $M^{pre}$ based on the image $I$ and instruction $Q$. In contrast, our method unifies these two tasks into a single framework, $(I,Q)\rightarrow (A^{pre},M^{pre})$, enabling simultaneous instruction-guided segmentation, open-ended VQA, and multiple-choice VQA. 
%
%

\textbf{GRPO for Visual Perception.}
Visual-RFT~\cite{liu2025visualrft} and Seg-Zero~\cite{liu2025segzero} extend the GRPO~\cite{shao2024deepseekmath} framework to visual perception tasks by introducing task-specific rewards. Given an input image $I$ and instruction $Q$, the model generates $n$ candidate predictions of the expected output, each of which is compared against ground-truth coordinates to compute individual rewards. GRPO then performs group-wise normalization over these rewards, guiding the model to favor perceptually accurate outputs, even in the absence of reasoning data during cold-start training. This approach enables more effective visual alignment and improves the generalization capability of MLLMs for object detection, classification, and referring segmentation. 

\begin{figure}[t]
  \centering
  \includegraphics[width=1\linewidth]{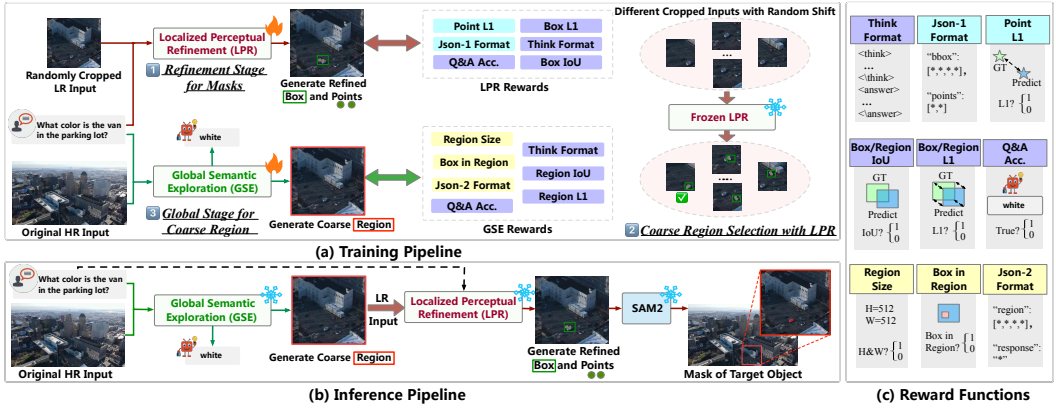}
  \caption{Framework overview of \textsc{FineRS}. (a) During training, we design specific reward functions to train GSE and LPR, where the LPR is optimized first and adopted to form a retrospective reward to enhance the coarse region accuracy of GSE. (b) During inference, GSE takes a high-resolution image and user instructions as input and produces an answer and a coarse region containing referred objects. Then, LPR processes the instruction and coarse region to generate the object box and adopts SAM2~\cite{ravi2024sam2} to generate the final mask. (c) To unify VQA and segmentation into a single MLLM, we design a multi-task reward pool and assign the items to supervise GSE and LPR. }
  
  \label{fig:framework}
  \vspace{-4mm}
\end{figure}

\subsection{Overall Framework of \textsc{FineRS}}

The overall framework of \textsc{FineRS} is illustrated in Fig.~\ref{fig:framework}. It consists of a two-stage MLLM pipeline, comprising Global Semantic Exploration (GSE) and Localized Perceptual Refinement (LPR). Unlike prior high-resolution MLLMs~\cite{wang2025divide,shen2024zoomeye} that rely on complex search strategies across multiple cropped image patches, our framework is designed to produce both textual responses and object segmentation masks through a single feedforward pass. 
%
As shown in Fig.~\ref{fig:framework} (a), during training, GSE and LPR are optimized independently with specially designed rewards to facilitate perception at different levels of granularity. We introduce a locate-informed retrospective reward, which leverages LPR to select a robust coarse region, enhancing the global exploration precision of GSE. The detailed reward formulations for both modules are shown in Fig.~\ref{fig:framework} (c).
During inference, as shown in Fig.~\ref{fig:framework} (b), only the original high-resolution image and user instruction are fed into GSE, which directly outputs the coarse box $B_r^{pre}$ and the text response $A^{pre}$. The GSE stage can be formulated as:
\begin{align}
(B_r^{pre},  A^{pre})= \mathcal{G}(\theta_{\texttt{GSE}}(I,Q)),
\end{align}
where $\mathcal{G}$ represents a post-processing function to extract the keywords of the long response. To constrain the search space of $B_r^{pre}$, we set a fixed box size to optimize only the center offsets. 
The coarse box indicates an enlarged region around the referred object, which will subsequently be passed to LPR for object localization and mask generation.
In the second stage, LPR first crops the original image based on $B_r^{pre}$ to obtain a lower-resolution input $I_c$, and performs local reasoning to generate bounding boxes $B^{pre}$ and points $P_1^{pre}, P_2^{pre}$. The LPR stage can be formulated as:
\begin{align}
(B^{pre},  P_1^{pre}, P_2^{pre})= \mathcal{G}(\theta_{\texttt{LPR}}(I_c,Q)).
\label{eq:LPR}
\end{align}
Finally, the generated boxes and points are passed to the frozen SAM2~\cite{ravi2024sam2} to produce the segmentation mask $M^{pre}$ for the target object.
%


\subsection{Training Pipeline}

The comparison in Tab.~\ref{tab:bench_compare} illustrates that the number of samples in existing high-resolution benchmarks is substantially lower than that of standard-resolution datasets. To avoid data overload of supervised fine-tuning, we exploit a recent vision reinforcement learning strategy to enhance the reasoning capacities of two-stage MLLMs. The training process is illustrated in Fig.~\ref{fig:framework} (a). 
Specifically, we train our \textsc{FineRS} model with the GRPO algorithm in two stages. First, the coarse-grained LPR module is trained with LPR rewards to generate the object box from a local image crop. Then, the GSE module is optimized using a combination of GSE rewards and a locate-informed retrospective reward guided by the LPR module. In addition, a standard KL divergence penalty is applied between the policy and the reference model.

\textbf{\ding{182} Refinement Stage for Masks.}
The supervised annotations for LPR training include bounding boxes $B^{gt}$, points $P_1^{gt}$ and $P_2^{gt}$, and text responses $A^{gt}$, all derived from the masks and answer annotations provided in training data. The <think> process originates from a prompt-based cold start and is supervised using the <think> format. Notably, since LPR is designed to focus on fine-grained perception within lower-resolution regions, it is trained on image patches that are randomly cropped around the ground-truth bounding boxes.

\textbf{\ding{183} Coarse Region Selection with LPR.}
Unlike LPR, the coarse region of GSE lacks explicit ground truth supervision for reward calculation. To address this, we design a locate-informed retrospective reward that uses the outputs of LPR to provide robust coarse region supervision for GSE. Specifically, for each training sample, we first generate $n$ randomly offset coarse regions that cover the GT bounding box $B^{gt}$. We then compute the IoU score between the LPR-predicted boxes $B^{pre}$ and the ground-truth boxes, selecting the region with the highest IoU as the GT regions $B_r^{gt}$ for training the GSE module. 

\textbf{\ding{184} Global Stage for Coarse Region.}  
Due to the complex scenes in high-resolution images, which make it challenging for the model to focus on small targets, the GSE model is designed to generate approximate regions where small targets are likely to exist, based on the instruction context. Therefore, the training data for GSE consists of high-resolution images annotated with the optimal regions $B_r^{gt}$ selected by LPR and corresponding ground-truth answer labels $A^{gt}$. 

\subsection{Reward Functions}
Inspired by the reward functions of the GRPO strategy, we design distinct reward functions at different levels of granularity for the LPR and GSE modules.

\textbf{Rewards for LPR.} The rewards for LPR include Point L1 $R_{point}$, Box L1 $R_{b^{L1}}$, Box IoU $R_{b^{IoU}}$, JSON-1 format $R_{format1}$, Think format $R_{think}$, and Q\&A Accuracy $R_{response}$. Among these functions, $R_{point}$, $R_{b^{L1}}$, and $R_{b^{IoU}}$ are computed based on the predicted boxes $B^{pre}$, predicted points $P^{pre}$, GT boxes $B^{gt}$ and GT points $P^{gt}$, to enforce spatial alignment between predictions and ground-truth annotations. The JSON-1 format reward $R_{format1}$ is only considered correct if the model outputs exact keywords \{bbox, points 1, points 2, response\} in the required structure. 
The response reward $R_{response}$ for the final response $A^{pre}$ is defined as:
\begin{align}
R_{response} = \left\{
\begin{array}{ll}
1 & ,\quad \text{if}~~  A^{pre} ~~ \text{is} ~~ \text{True},\\
0 & ,\quad \text{if} ~~  A^{pre}~~ \text{is} ~~ \text{False}, \\
\end{array}
\right.
\end{align}
where the criteria for determining whether $A^{pre}$ is correct vary across task settings. For instruction-guided segmentation, the response is considered correct if it includes phrases like “is detected” or “is found”. In the multiple-choice VQA setting, the response is correct if it exactly matches the ground-truth option. In the open-ended VQA setting, the response is deemed correct if the fuzzy matching similarity to the ground-truth answer exceeds 0.8. The final reward of LPR is computed as:
\begin{align}
R_{LPR} = R_{b^{IoU}} + R_{b^{L1}} + R_{point} + R_{format1} + R_{response} + R_{think}.
\label{eq:RLPR}
\end{align}

\textbf{Rewards for GSE.} Unlike LPR, the reward functions for GSE are designed to encourage alignment between the predicted coarse region $B_r^{pre}$ and the ground-truth region $B_r^{gt}$ selected by LPR. Specifically, we name this reward as locate-informed retrospective reward that consists of a region IoU reward $R_{region^{IoU}}$ and a region L1 $R_{region^{L1}}$ between the predicted and GT regions ${B_r^{pre}, B_r^{gt}}$. Since this stage focuses solely on contextual regions rather than fine-grained localization, the point-level reward is omitted, and the output JSON format is updated to a new template as \{region, response\}.
Additionally, to ensure that the predicted regions are compatible with the input size expected by the LPR module and sufficiently cover the target object, we introduce a region size $R_{size}$ reward and a box-in-region $R_{cover}$ reward. They encourage the model to generate regions of appropriate size and position, aligned with those used during LPR training. The think format and Q\&A accuracy rewards are kept consistent with those used in LPR. The final reward of GSE is computed as:
\begin{align}
R_{GSE} = R_{region^{IoU}} + R_{region^{L1}} + R_{size} + R_{cover} + R_{format2} + R_{response} + R_{think}.
\label{eq:RGSE}
\end{align}

\begin{table*}[t]
\scriptsize
\centering
\caption{Performance comparison on the test set of \textsc{FineRS}-4k. ``$\dagger$'' indicates that the corresponding method is retrained with our dataset. We label the best results with a \textbf{bold} style.}
\begin{tabular}{|l||
    >{\centering\arraybackslash}p{0.6cm} 
    >{\centering\arraybackslash}p{0.6cm} 
    >{\centering\arraybackslash}p{0.6cm}
    >{\centering\arraybackslash}p{0.8cm}|
    >{\centering\arraybackslash}p{0.6cm}
    >{\centering\arraybackslash}p{0.6cm} 
    >{\centering\arraybackslash}p{0.6cm}
    >{\centering\arraybackslash}p{0.6cm}
    >{\centering\arraybackslash}p{0.8cm}|
}
\hline
\rowcolor{gray!45}   & \multicolumn{4}{c|}{IoU (gIoU/cIoU)} &  \multicolumn{5}{c|}{QA Acc. (MVQA/OVQA)} \\ 

\rowcolor{gray!45} ~~~~~~~~~~~~~~~~~~~~~\multirow{-2}{*}{Method}  & \textit{S} & \textit{xS} & \textit{xxS} & \textit{All} & \textit{Color} & \textit{Shape} & \textit{Others} & \textit{Position} & \textit{All} \\
\hline
\hline
\rowcolor{gray!10}
\multicolumn{10}{|l|}{\textit{Training-free}} \\

LISA \tiny \texttt{7B}~\cite{lai2024lisa} &19.1/6.49 &8.28/1.20&4.19/0.34&9.00/2.38 &0.00/6.11&0.00/0.00&0.00/9.37&0.00/16.7&0.00/5.51\\
LISA \tiny \texttt{13B}~\cite{lai2024lisa} & 16.4/3.86 & 7.02/0.73 & 2.55/0.18 & 7.29/1.42 &0.00/6.46&0.00/6.55&0.00/9.37&0.00/5.55&0.00/6.58\\
LISA++ \tiny \texttt{7B}~\cite{yang2023lisa++} & 25.9/12.3&13.5/2.90&3.70/0.70&12.3/5.20&5.90/9.79&0.82/1.63&35.7/6.24&26.3/5.55&6.72/8.19 \\
PixelLM \tiny \texttt{7B}~\cite{ren2024pixellm} & 13.6/6.70&3.30/1.10&0.50/0.10&4.40/2.10&0.00/4.56&0.82/0.09&3.57/9.31&0.09/5.55&0.27/4.05\\
SEAL~\cite{wu2024v} &-- & -- &--  & -- & 7.46/3.14&8.26/0.03&7.14/15.6&5.26/16.7&7.53/3.49      \\
DC$^2$ ~\cite{wang2025divide} &-- & -- &--  & -- & 39.0/19.5 &36.3/9.00&57.1/21.8&36.8/16.6&39.2/17.8 \\ 
MLLMs-Know \tiny \texttt{7B}~\cite{zhang2025mllms}  &-- & -- &--  & -- & 52.8/50.2&{45.5}/{32.0}& 60.7/34.4& 36.8/5.56& 51.5/45.4  \\
MLLMs-Know \tiny \texttt{13B}~\cite{zhang2025mllms} &-- & -- &--  & -- &56.1/54.0&36.4/32.0&67.9/37.5 &26.3/16.7&52.6/48.8    \\
MLLMs-Know \tiny{\texttt{7B}~\cite{zhang2025mllms}} + \scriptsize LISA \tiny \texttt{7B}~\cite{lai2024lisa} & 21.2/19.2& 15.0/7.95& 11.3/4.50& 14.9/12.5& 52.8/50.2& 45.5/32.0& 60.7/34.4& 36.8/5.56& 51.5/45.4  \\
MLLMs-Know \tiny \texttt{13B}~\cite{zhang2025mllms} + \scriptsize LISA \tiny \texttt{13B}~\cite{lai2024lisa} & 27.0/20.3& 18.3/9.29 & 12.4/3.62&  17.9/12.8& {56.1}/{54.0}&36.4/32.0&67.9/37.5 &26.3/16.7&52.6/48.8    \\
Seg-Zero \tiny \texttt{7B}~\cite{liu2025segzero} & 55.9/18.6 & 34.7/4.94 & 16.5/0.84 & 32.1/6.61 & -- & -- & -- & -- & --\\
\hline

\rowcolor{gray!10}
\multicolumn{10}{|l|}{\textit{Training}} \\
LISA$^\dagger$ \tiny \texttt{7B}~\cite{lai2024lisa} & 13.0/9.62&15.0/11.4&8.78/5.59&12.1/9.64 & 23.9/24.3&24.8/18.6&25.0/6.2&21.1/0.0&23.9/22.0\\
PixelLM$^\dagger$ \tiny \texttt{7B}~\cite{ren2024pixellm} & 1.27/1.02&
0.52/0.35&
0.08/0.02&
0.16/0.13&
0.0/0.0&
0.0/0.0&
0.0/0.0& 
0.0/0.0& 
0.0/0.0 \\
MLLMs-Know \tiny{\texttt{7B}~\cite{zhang2025mllms}} + \scriptsize LISA$^\dagger$ \tiny \texttt{7B}~\cite{lai2024lisa}  & 1.32/1.16 &2.08/1.90&2.85/3.15&2.22/1.57  & 52.8/50.2& 45.5/32.0& 60.7/34.4& 36.8/5.56& 51.5/45.4\\
Seg-zero$^\dagger$ \tiny \texttt{7B}~\cite{liu2025segzero} & {61.8}/{50.5} & {53.0}/{30.2} & {31.7}/{20.7} & {46.6}/{38.6} & -- & -- & -- & -- & --\\
\hline
\hline
\textbf{Ours (\textsc{FineRS})} \tiny \texttt{7B} & \textbf{62.2/52.6} & \textbf{59.0/43.1} & \textbf{47.2/27.5} & \textbf{55.1/46.5} & \textbf{85.8/60.5} & \textbf{76.0/49.2} & \textbf{78.6/34.4}& \textbf{63.2/27.8} & \textbf{83.3/56.7}\\
\hline
\end{tabular}
\label{tab:method_compare}
\end{table*}

\section{Experiments}
\subsection{Experimental Settings}
\textbf{Implementation Details.}
Our two-stage MLLMs are built upon Qwen2.5-VL-7B~\cite{yang2024qwen2}, with input resolution of $1920\times1080$ for GSE and $512\times512$ for LPR. The output coarse region of GSE is $256\times256$, which will be $2\times$ upsampled before being fed into LPR. The whole model is trained on a 4$\times$A800 GPU (80G) setup using the Seg-Zero~\cite{liu2025segzero} and DeepSpeed~\cite{guo2025deepseek} library. During training, the GSE module uses a total batch size of 16 with 8 samples per training step, while the LPR module uses a total batch size of 32, also with 8 samples per step. For both stages, the initial learning rate is set to 1e-6 and the weight decay is 0.01. In addition, we adopt SAM2~\cite{ravi2024sam2} for box-to-mask generation, which is kept frozen during training. The user prompts for GSE and LPR across three tasks are presented in Fig.~\ref{Ap:GSEprompt} and Fig.~\ref{Ap:LPRprompt}.

\textbf{Evaluation Metrics.}
Following previous works~\cite{yu2016modeling, kazemzadeh2014referitgame}, we calculate gIoU and cIoU for instruction-guided segmentation. The gIoU is the average of all per-image Intersection-over-Unions (IoUs), while the cIoU calculates the cumulative intersection over the cumulative union. We evaluate cIoU and gIoU metrics across different object sizes. In addition, we calculate the accuracy of multiple-choice (MVQA) and open-ended (OVQA) visual question answering using option accuracy and the ``difflib.SequenceMatcher'' algorithm, with a matching threshold set to 0.8. 

\subsection{Comparison with State-of-the-art Methods}
We evaluate \textsc{FineRS} and other methods on our \textsc{FineRS} dataset across three tasks, including instruction-guided segmentation, open-ended VQA, and multiple-choice VQA. 

\textbf{\ding{182} Comparison on \textsc{FineRS}-4k.} The comparison results on the test and validation sets of our \textsc{FineRS}-4k are illustrated in Tab.~\ref{tab:method_compare} and Tab.~\ref{Aptab:2}, respectively. We report both training-free approaches and selected retrained methods in our dataset. As shown, our model consistently outperforms state-of-the-art segmentation approaches and high-resolution VQA methods. Note that the IoU scores are computed on all samples from the three instruction types, while QA accuracy is calculated only on the samples of MVQA and OVQA. 
The performance drop of LISA~\cite{lai2024lisa} and PixelLM~\cite{ren2024pixellm} likely arises from the domain shift and task complexity of \textsc{FineRS}-4k, which contains 4K UAV imagery with ultra-small, sparse objects. All baselines are fine-tuned on \textsc{FineRS}-4k under the same settings. In this low-data, high-resolution regime, fixed-architecture models tend to overfit or underfit.

\textbf{\ding{183} Comparison on Other VQA Datasets.} We also conduct a comparison with other approaches on public high-resolution VQA datasets, including V*~\cite{wu2024v} and HR-Bench~\cite{wang2025divide}. Compared with our benchmark, these datasets are captured in a general, non-UVA perception with optional VAQ annotations. As shown in Tab.~\ref{tab:method_compare2}, without an additional finetuning process, our model achieves significantly better accuracy in non-UVA scenarios from 4k to 8k resolutions. Note that SEAL~\cite{wu2024v} is the model proposed in the V* benchmark. 

\begin{figure}[t]
  \centering
  \includegraphics[width=0.95\linewidth]{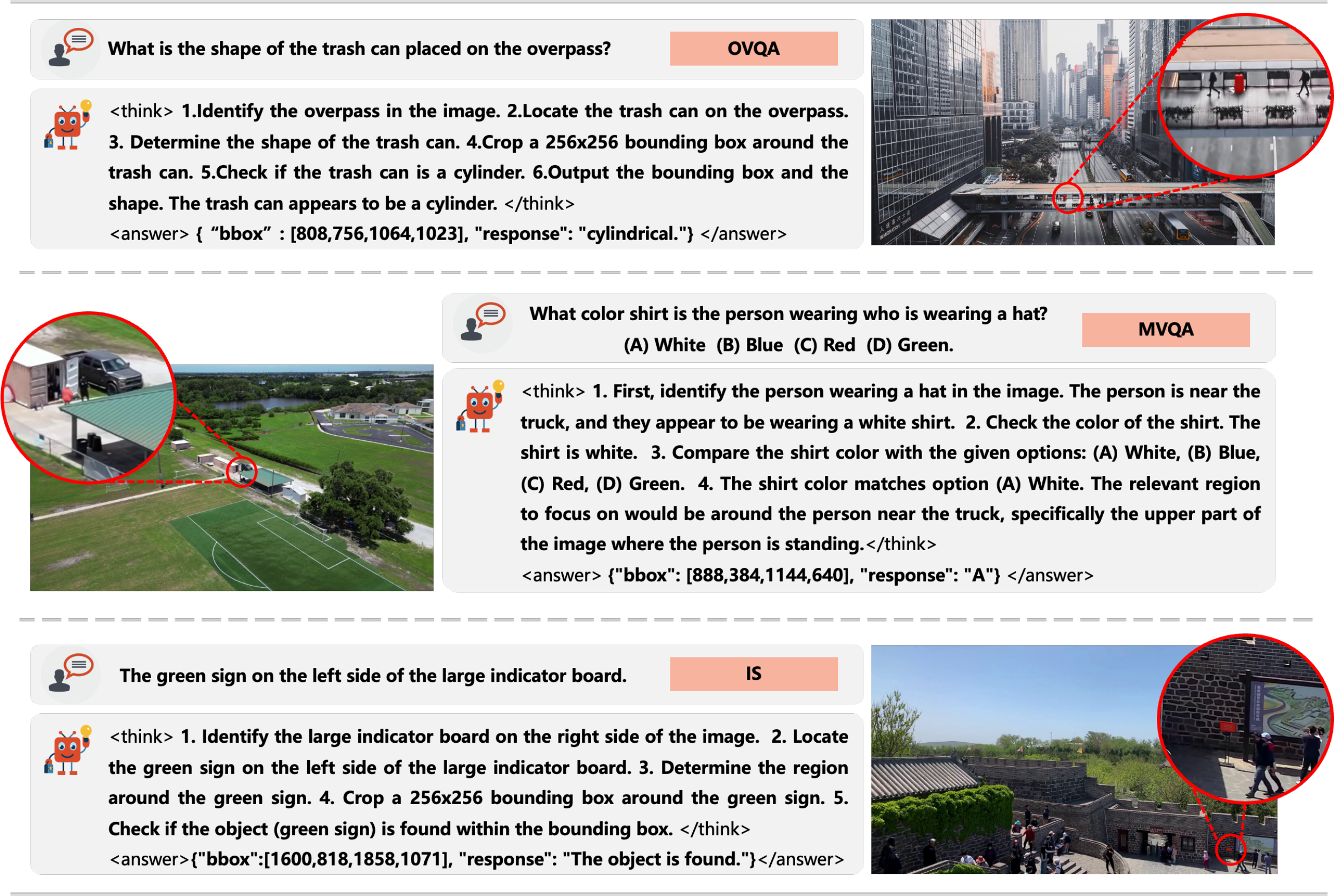}
  \caption{Visual results on Open-ended VQA (OVQA), Multiple-choice VQA (MVQA), and Instruction-guided Segmentation (IS). The listed images are sampled from \textsc{FineRS}-4k test set. }

  \label{fig:visualization}
\end{figure}

\textbf{\ding{184} Visualization Results.} Representative visual results for all three tasks are provided in Fig.~\ref{fig:visualization}. As shown, the proposed two-stage framework is effective in locating tiny objects from a cluttered background and generating accurate open-ended or optional answers from the textual instructions.

\begin{table*}[t]
\scriptsize
\centering

\caption{Performance comparison on other high-resolution VQA benchmarks. ``\textit{Attr.}'' and ``\textit{Spat.}'' denote attribute and spatial, respectively. We label the best methods with a \textbf{bold} style.}
\begin{tabular}{|l|| 
    >{\centering\arraybackslash}p{1.3cm}|
    >{\centering\arraybackslash}p{0.53cm} 
    >{\centering\arraybackslash}p{0.53cm} 
    >{\centering\arraybackslash}p{0.63cm}|
    >{\centering\arraybackslash}p{0.53cm} 
    >{\centering\arraybackslash}p{0.53cm} 
    >{\centering\arraybackslash}p{0.53cm}|
    >{\centering\arraybackslash}p{0.53cm} 
    >{\centering\arraybackslash}p{0.53cm} 
    >{\centering\arraybackslash}p{0.53cm}|
}
\hline
\rowcolor{gray!45} ~ &  & \multicolumn{3}{c|}{V$^*$} &  \multicolumn{3}{c|}{HR-Bench 4K} &  \multicolumn{3}{c|}{HR-Bench 8K}\\ 
\rowcolor{gray!45} ~~~~~~~~~~~~~ \multirow{-2}{*}{Method} ~~~~~~~~~ & \multirow{-2}{*}{Segmentation} & \textit{Attr.} & \textit{Spat.} & \textit{Overall} & \textit{FSP} & \textit{FCP} & \textit{Avg.} & \textit{FSP} & \textit{FCP} & \textit{Avg.}  \\
\hline
\hline

SEAL~\cite{wu2024v} & \xmark  &{74.8} &{76.3} &{75.4}&  47.0&29.3&38.1&  42.5&28.8&35.6     \\
Mllms-Know~\cite{zhang2025mllms}& \xmark & -- & -- & 62.3& 52.4&30.2&41.3& {47.2} &30.7&38.9 \\
VILA-HD-1.5K~\cite{shi2025scaling}& \xmark  & -- & -- & 68.1& -- & -- &--& -- & -- &--  \\
VILA-HD-4K~\cite{shi2025scaling}& \xmark &  -- & -- & 71.2 & -- & -- &--&--  &--  & --\\
DC$^2$~\cite{wang2025divide} & \xmark &  -- & -- & 57.3 &  {53.0} &{47.0} &{50.0}& 37.2&{44.2}&{40.8}    \\
\hline
\hline
\textbf{Ours (\textsc{FineRS})} & \cmark &  \textbf{76.5} & \textbf{79.0}& \textbf{77.5}& \textbf{66.4}&
\textbf{61.2}&\textbf{63.8}  & \textbf{60.2}&\textbf{55.9}&\textbf{58.1}     \\
\hline

\end{tabular}
\label{tab:method_compare2}
\end{table*}

\subsection{Ablation Study}


Tab.~\ref{tab:model_evaluation} illustrates the evaluation results by separately removing the key components from the whole framework. To validate the efficacy of retrospective reward, we replace the LPR-informed coarse regions with a random crop centered around the GT object to supervise the GSE for coarse box generation. The result of ``\textit{w/o Restrospective Reward}'' demonstrates the effectiveness of this design. 
To mitigate the sensitivity of LPR to input box variations, we apply a box augmentation strategy during LPR training. The improvement observed in ``\textit{w/o Random Input Region in LPR}'' verifies the efficacy of this data augmentation strategy. 
In contrast to \cite{liu2025segzero}, our \textsc{FineRS} can generate exact text answers to user questions, which requires the incorporation of a QA accuracy reward during training. The performance drop in ``\textit{w/o QA Acc. Reward}'' supports the importance of this component. More detailed analysis on this reward can be found in Tab.~\ref{ablation2}.
In GSE, we use extra box rewards to facilitate the optimization of coarse region generation. The ablation results of ``\textit{w/o Box Size Reward}'' and ``\textit{w/o Box-in-region Reward}'' demonstrate the positive impact of these rewards on model performance.

\begin{table*}[t]
\scriptsize
\centering

\caption{Ablation studies about the proposed components on \textsc{FineRS}-4k.}
\begin{tabular}{|l||
    >{\centering\arraybackslash}p{0.6cm} 
    >{\centering\arraybackslash}p{0.7cm}
    >{\centering\arraybackslash}p{0.6cm} 
    >{\centering\arraybackslash}p{0.6cm}|
    >{\centering\arraybackslash}p{0.7cm}
    >{\centering\arraybackslash}p{0.6cm} 
    >{\centering\arraybackslash}p{0.6cm} 
    >{\centering\arraybackslash}p{0.7cm}|
}
\hline
\rowcolor{gray!45} ~ & \multicolumn{4}{c|}{Test set} & \multicolumn{4}{c|}{Val set} \\ 
\rowcolor{gray!45} ~~~~~~~~~~~~~~~~~~~~~~\multirow{-2}{*}{Different Settings} ~~~~~~~~~~~~~~~~~~~~~~~ & \textit{gIoU} & \textit{cIoU} & \textit{MVQA} & \textit{OVQA} & \textit{gIoU} & \textit{cIoU} & \textit{MVQA} & \textit{OVQA}  \\
\hline
\hline

\textbf{\textsc{FineRS}} & \textbf{55.1} & \textbf{46.5}&\textbf{83.3} & \textbf{56.7}&\textbf{49.9}&  \textbf{39.4} &\textbf{87.2}&\textbf{60.0}  \\
~~~~~~ \textit{w/o Retrospective Reward}&  54.0& 44.0 & 82.3 & 53.0 & 49.4  & 38.0 & 86.2 & 55.8  \\
~~~~~~ \textit{w/o  Random Input Region in LPR} & 53.9 & 46.7 &  83.7&61.9 & 48.7& 39.4 &84.8  & 59.4  \\
~~~~~~ \textit{w/o  QA Acc. Reward}& 52.8 & 45.7  &  --& -- & 48.6& 33.7 & -- &  --  \\
~~~~~~ \textit{w/o Box Size Reward}&  51.0 &43.4 & 56.5 & 40.8& 44.5& 35.8 &78.4  &  56.4     \\
~~~~~~ \textit{w/o Box-in-region Reward}& 50.1 & 42.0 & 56.5 & 40.8 & 44.1 & 34.2 & 77.9 &55.7    \\

\hline
\end{tabular}
\label{tab:model_evaluation}
\vspace{-4mm}
\end{table*}

\section{Discussion}
In this paper, we aim to resolve the fine-grained reasoning and segmentation of ultra-small objects in high-resolution images. We introduce 1) \textsc{FineRS}, a two-stage MLLM-based reinforcement learning framework that combines global semantic exploration with localized perceptual refinement; and 2) \textsc{FineRS}-4k, a new dataset featuring challenging scenes annotated with text-mask pairs across three types of tasks. Extensive experiments on \textsc{FineRS}-4k and other public benchmarks demonstrate the superiority of the proposed method in both answering accuracy and segmentation precision.  

Although our method made an early attempt to apply reinforcement learning for joint object reasoning and segmentation, several limitations remain in the current architecture. First, the localization accuracy of LPR is highly dependent on the output coarse region of GSE. The text answer and object box will be incorrect if the GT object exceeds the coarse region. Incorporating an additional signal for object existence and enabling re-exploration could help mitigate such missed detection errors. Second, due to memory constraints, the two-stage models are not jointly optimized during training. Designing a more efficient architecture for jointly end-to-end training of two stages remains an important future direction.   

\section*{Acknowledgment}
This work was supported by National Natural Science Foundation of China under Grant 62206039, 62406053, 62476048, 62293544, the Fundamental Research Funds for the Central Universities under DUT24RC(3)025, DUT24YG119, Sichuan Science and Technology Program (2025YFMS0004), and a part by 2024-0011 (ZX20240867).
{
    \small
    \bibliographystyle{unsrt}
    \bibliography{ref}
}
\medskip

\newpage

\clearpage
\section*{NeurIPS Paper Checklist}

\begin{enumerate}

\item {\bf Claims}
    \item[] Question: Do the main claims made in the abstract and introduction accurately reflect the paper's contributions and scope?
    \item[] Answer: \answerYes 
    \item[] Justification: The motivations and contributions of this paper is highly summarized in our abstract and introduction.
    \item[] Guidelines:
    \begin{itemize}
        \item The answer NA means that the abstract and introduction do not include the claims made in the paper.
        \item The abstract and/or introduction should clearly state the claims made, including the contributions made in the paper and important assumptions and limitations. A No or NA answer to this question will not be perceived well by the reviewers. 
        \item The claims made should match theoretical and experimental results, and reflect how much the results can be expected to generalize to other settings. 
        \item It is fine to include aspirational goals as motivation as long as it is clear that these goals are not attained by the paper. 
    \end{itemize}

\item {\bf Limitations}
    \item[] Question: Does the paper discuss the limitations of the work performed by the authors?
    \item[] Answer: \answerYes 
    \item[] Justification: We discussed the limitations in the conclusion section.
    \item[] Guidelines:
    \begin{itemize}
        \item The answer NA means that the paper has no limitation while the answer No means that the paper has limitations, but those are not discussed in the paper. 
        \item The authors are encouraged to create a separate "Limitations" section in their paper.
        \item The paper should point out any strong assumptions and how robust the results are to violations of these assumptions (e.g., independence assumptions, noiseless settings, model well-specification, asymptotic approximations only holding locally). The authors should reflect on how these assumptions might be violated in practice and what the implications would be.
        \item The authors should reflect on the scope of the claims made, e.g., if the approach was only tested on a few datasets or with a few runs. In general, empirical results often depend on implicit assumptions, which should be articulated.
        \item The authors should reflect on the factors that influence the performance of the approach. For example, a facial recognition algorithm may perform poorly when image resolution is low or images are taken in low lighting. Or a speech-to-text system might not be used reliably to provide closed captions for online lectures because it fails to handle technical jargon.
        \item The authors should discuss the computational efficiency of the proposed algorithms and how they scale with dataset size.
        \item If applicable, the authors should discuss possible limitations of their approach to address problems of privacy and fairness.
        \item While the authors might fear that complete honesty about limitations might be used by reviewers as grounds for rejection, a worse outcome might be that reviewers discover limitations that aren't acknowledged in the paper. The authors should use their best judgment and recognize that individual actions in favor of transparency play an important role in developing norms that preserve the integrity of the community. Reviewers will be specifically instructed to not penalize honesty concerning limitations.
    \end{itemize}

\item {\bf Theory assumptions and proofs}
    \item[] Question: For each theoretical result, does the paper provide the full set of assumptions and a complete (and correct) proof?
    \item[] Answer: \answerNA{} 
    \item[] Justification: This paper focuses on the application of MLLMs without theoretical components.
    \item[] Guidelines:
    \begin{itemize}
        \item The answer NA means that the paper does not include theoretical results. 
        \item All the theorems, formulas, and proofs in the paper should be numbered and cross-referenced.
        \item All assumptions should be clearly stated or referenced in the statement of any theorems.
        \item The proofs can either appear in the main paper or the supplemental material, but if they appear in the supplemental material, the authors are encouraged to provide a short proof sketch to provide intuition. 
        \item Inversely, any informal proof provided in the core of the paper should be complemented by formal proofs provided in appendix or supplemental material.
        \item Theorems and Lemmas that the proof relies upon should be properly referenced. 
    \end{itemize}

    \item {\bf Experimental result reproducibility}
    \item[] Question: Does the paper fully disclose all the information needed to reproduce the main experimental results of the paper to the extent that it affects the main claims and/or conclusions of the paper (regardless of whether the code and data are provided or not)?
    \item[] Answer: \answerYes{} 
    \item[] Justification: We have included the implementation details and will provide the code, models, and dataset once publication.
    \item[] Guidelines:
    \begin{itemize}
        \item The answer NA means that the paper does not include experiments.
        \item If the paper includes experiments, a No answer to this question will not be perceived well by the reviewers: Making the paper reproducible is important, regardless of whether the code and data are provided or not.
        \item If the contribution is a dataset and/or model, the authors should describe the steps taken to make their results reproducible or verifiable. 
        \item Depending on the contribution, reproducibility can be accomplished in various ways. For example, if the contribution is a novel architecture, describing the architecture fully might suffice, or if the contribution is a specific model and empirical evaluation, it may be necessary to either make it possible for others to replicate the model with the same dataset, or provide access to the model. In general. releasing code and data is often one good way to accomplish this, but reproducibility can also be provided via detailed instructions for how to replicate the results, access to a hosted model (e.g., in the case of a large language model), releasing of a model checkpoint, or other means that are appropriate to the research performed.
        \item While NeurIPS does not require releasing code, the conference does require all submissions to provide some reasonable avenue for reproducibility, which may depend on the nature of the contribution. For example
        \begin{enumerate}
            \item If the contribution is primarily a new algorithm, the paper should make it clear how to reproduce that algorithm.
            \item If the contribution is primarily a new model architecture, the paper should describe the architecture clearly and fully.
            \item If the contribution is a new model (e.g., a large language model), then there should either be a way to access this model for reproducing the results or a way to reproduce the model (e.g., with an open-source dataset or instructions for how to construct the dataset).
            \item We recognize that reproducibility may be tricky in some cases, in which case authors are welcome to describe the particular way they provide for reproducibility. In the case of closed-source models, it may be that access to the model is limited in some way (e.g., to registered users), but it should be possible for other researchers to have some path to reproducing or verifying the results.
        \end{enumerate}
    \end{itemize}

\item {\bf Open access to data and code}
    \item[] Question: Does the paper provide open access to the data and code, with sufficient instructions to faithfully reproduce the main experimental results, as described in supplemental material?
    \item[] Answer: \answerYes{} 
    \item[] Justification: We promise to release the code and data after acceptance. 
    \item[] Guidelines:
    \item[] Guidelines:
    \begin{itemize}
        \item The answer NA means that paper does not include experiments requiring code.
        \item Please see the NeurIPS code and data submission guidelines (\url{https://nips.cc/public/guides/CodeSubmissionPolicy}) for more details.
        \item While we encourage the release of code and data, we understand that this might not be possible, so “No” is an acceptable answer. Papers cannot be rejected simply for not including code, unless this is central to the contribution (e.g., for a new open-source benchmark).
        \item The instructions should contain the exact command and environment needed to run to reproduce the results. See the NeurIPS code and data submission guidelines (\url{https://nips.cc/public/guides/CodeSubmissionPolicy}) for more details.
        \item The authors should provide instructions on data access and preparation, including how to access the raw data, preprocessed data, intermediate data, and generated data, etc.
        \item The authors should provide scripts to reproduce all experimental results for the new proposed method and baselines. If only a subset of experiments are reproducible, they should state which ones are omitted from the script and why.
        \item At submission time, to preserve anonymity, the authors should release anonymized versions (if applicable).
        \item Providing as much information as possible in supplemental material (appended to the paper) is recommended, but including URLs to data and code is permitted.
    \end{itemize}

\item {\bf Experimental setting/details}
    \item[] Question: Does the paper specify all the training and test details (e.g., data splits, hyperparameters, how they were chosen, type of optimizer, etc.) necessary to understand the results?
    \item[] Answer: \answerYes{} 
    \item[] Justification: We clarify the experiment details in the experiment section. 
    \item[] Guidelines:
    \begin{itemize}
        \item The answer NA means that the paper does not include experiments.
        \item The experimental setting should be presented in the core of the paper to a level of detail that is necessary to appreciate the results and make sense of them.
        \item The full details can be provided either with the code, in appendix, or as supplemental material.
    \end{itemize}

\item {\bf Experiment statistical significance}
    \item[] Question: Does the paper report error bars suitably and correctly defined or other appropriate information about the statistical significance of the experiments?
    \item[] Answer: \answerYes{} 
    \item[] Justification: We have discussed the metrics in experiment section.
    \item[] Guidelines:
    \begin{itemize}
        \item The answer NA means that the paper does not include experiments.
        \item The authors should answer "Yes" if the results are accompanied by error bars, confidence intervals, or statistical significance tests, at least for the experiments that support the main claims of the paper.
        \item The factors of variability that the error bars are capturing should be clearly stated (for example, train/test split, initialization, random drawing of some parameter, or overall run with given experimental conditions).
        \item The method for calculating the error bars should be explained (closed form formula, call to a library function, bootstrap, etc.)
        \item The assumptions made should be given (e.g., Normally distributed errors).
        \item It should be clear whether the error bar is the standard deviation or the standard error of the mean.
        \item It is OK to report 1-sigma error bars, but one should state it. The authors should preferably report a 2-sigma error bar than state that they have a 96\% CI, if the hypothesis of Normality of errors is not verified.
        \item For asymmetric distributions, the authors should be careful not to show in tables or figures symmetric error bars that would yield results that are out of range (e.g. negative error rates).
        \item If error bars are reported in tables or plots, The authors should explain in the text how they were calculated and reference the corresponding figures or tables in the text.
    \end{itemize}

\item {\bf Experiments compute resources}
    \item[] Question: For each experiment, does the paper provide sufficient information on the computer resources (type of compute workers, memory, time of execution) needed to reproduce the experiments?
    \item[] Answer: \answerYes{} 
    \item[] Justification: We have claimed the used computer resources.
    \item[] Guidelines:
    \begin{itemize}
        \item The answer NA means that the paper does not include experiments.
        \item The paper should indicate the type of compute workers CPU or GPU, internal cluster, or cloud provider, including relevant memory and storage.
        \item The paper should provide the amount of compute required for each of the individual experimental runs as well as estimate the total compute. 
        \item The paper should disclose whether the full research project required more compute than the experiments reported in the paper (e.g., preliminary or failed experiments that didn't make it into the paper). 
    \end{itemize}
    
\item {\bf Code of ethics}
    \item[] Question: Does the research conducted in the paper conform, in every respect, with the NeurIPS Code of Ethics \url{https://neurips.cc/public/EthicsGuidelines}?
    \item[] Answer: \answerYes{} 
    \item[] Justification: We have checked it.
    \item[] Guidelines:
    \begin{itemize}
        \item The answer NA means that the authors have not reviewed the NeurIPS Code of Ethics.
        \item If the authors answer No, they should explain the special circumstances that require a deviation from the Code of Ethics.
        \item The authors should make sure to preserve anonymity (e.g., if there is a special consideration due to laws or regulations in their jurisdiction).
    \end{itemize}

\item {\bf Broader impacts}
    \item[] Question: Does the paper discuss both potential positive societal impacts and negative societal impacts of the work performed?
    \item[] Answer: \answerNA{} 
    \item[] Justification: There is no societal impact of the work performed.
    \item[] Guidelines:
    \begin{itemize}
        \item The answer NA means that there is no societal impact of the work performed.
        \item If the authors answer NA or No, they should explain why their work has no societal impact or why the paper does not address societal impact.
        \item Examples of negative societal impacts include potential malicious or unintended uses (e.g., disinformation, generating fake profiles, surveillance), fairness considerations (e.g., deployment of technologies that could make decisions that unfairly impact specific groups), privacy considerations, and security considerations.
        \item The conference expects that many papers will be foundational research and not tied to particular applications, let alone deployments. However, if there is a direct path to any negative applications, the authors should point it out. For example, it is legitimate to point out that an improvement in the quality of generative models could be used to generate deepfakes for disinformation. On the other hand, it is not needed to point out that a generic algorithm for optimizing neural networks could enable people to train models that generate Deepfakes faster.
        \item The authors should consider possible harms that could arise when the technology is being used as intended and functioning correctly, harms that could arise when the technology is being used as intended but gives incorrect results, and harms following from (intentional or unintentional) misuse of the technology.
        \item If there are negative societal impacts, the authors could also discuss possible mitigation strategies (e.g., gated release of models, providing defenses in addition to attacks, mechanisms for monitoring misuse, mechanisms to monitor how a system learns from feedback over time, improving the efficiency and accessibility of ML).
    \end{itemize}
    
\item {\bf Safeguards}
    \item[] Question: Does the paper describe safeguards that have been put in place for responsible release of data or models that have a high risk for misuse (e.g., pretrained language models, image generators, or scraped datasets)?
    \item[] Answer: \answerYes{} 
    \item[] Justification: Our data comes from Internet videos or videos shot by ourselves, and there is no copyright dispute. We mosaic the sensitive data to prevent privacy leakage.
    \item[] Guidelines:
    \begin{itemize}
        \item The answer NA means that the paper poses no such risks.
        \item Released models that have a high risk for misuse or dual-use should be released with necessary safeguards to allow for controlled use of the model, for example by requiring that users adhere to usage guidelines or restrictions to access the model or implementing safety filters. 
        \item Datasets that have been scraped from the Internet could pose safety risks. The authors should describe how they avoided releasing unsafe images.
        \item We recognize that providing effective safeguards is challenging, and many papers do not require this, but we encourage authors to take this into account and make a best faith effort.
    \end{itemize}

\item {\bf Licenses for existing assets}
    \item[] Question: Are the creators or original owners of assets (e.g., code, data, models), used in the paper, properly credited and are the license and terms of use explicitly mentioned and properly respected?
    \item[] Answer: \answerYes{} 
    \item[] Justification: The baseline code used are open-source from Inertnet.
    \item[] Guidelines:
    \begin{itemize}
        \item The answer NA means that the paper does not use existing assets.
        \item The authors should cite the original paper that produced the code package or dataset.
        \item The authors should state which version of the asset is used and, if possible, include a URL.
        \item The name of the license (e.g., CC-BY 4.0) should be included for each asset.
        \item For scraped data from a particular source (e.g., website), the copyright and terms of service of that source should be provided.
        \item If assets are released, the license, copyright information, and terms of use in the package should be provided. For popular datasets, \url{paperswithcode.com/datasets} has curated licenses for some datasets. Their licensing guide can help determine the license of a dataset.
        \item For existing datasets that are re-packaged, both the original license and the license of the derived asset (if it has changed) should be provided.
        \item If this information is not available online, the authors are encouraged to reach out to the asset's creators.
    \end{itemize}

\item {\bf New assets}
    \item[] Question: Are new assets introduced in the paper well documented and is the documentation provided alongside the assets?
    \item[] Answer: \answerYes{} 
    \item[] Justification: We have processed the data to avoid this issue.
    \item[] Guidelines:
    \begin{itemize}
        \item The answer NA means that the paper does not release new assets.
        \item Researchers should communicate the details of the dataset/code/model as part of their submissions via structured templates. This includes details about training, license, limitations, etc. 
        \item The paper should discuss whether and how consent was obtained from people whose asset is used.
        \item At submission time, remember to anonymize your assets (if applicable). You can either create an anonymized URL or include an anonymized zip file.
    \end{itemize}

\item {\bf Crowdsourcing and research with human subjects}
    \item[] Question: For crowdsourcing experiments and research with human subjects, does the paper include the full text of instructions given to participants and screenshots, if applicable, as well as details about compensation (if any)? 
    \item[] Answer: \answerNA{} 
    \item[] Justification: The paper does not involve crowdsourcing nor research with human subjects.
    \item[] Guidelines:
    \begin{itemize}
        \item The answer NA means that the paper does not involve crowdsourcing nor research with human subjects.
        \item Including this information in the supplemental material is fine, but if the main contribution of the paper involves human subjects, then as much detail as possible should be included in the main paper. 
        \item According to the NeurIPS Code of Ethics, workers involved in data collection, curation, or other labor should be paid at least the minimum wage in the country of the data collector. 
    \end{itemize}

\item {\bf Institutional review board (IRB) approvals or equivalent for research with human subjects}
    \item[] Question: Does the paper describe potential risks incurred by study participants, whether such risks were disclosed to the subjects, and whether Institutional Review Board (IRB) approvals (or an equivalent approval/review based on the requirements of your country or institution) were obtained?
    \item[] Answer: \answerNA{} 
    \item[] Justification: The paper does not involve crowdsourcing nor research with human subjects.
    \item[] Guidelines:
    \begin{itemize}
        \item The answer NA means that the paper does not involve crowdsourcing nor research with human subjects.
        \item Depending on the country in which research is conducted, IRB approval (or equivalent) may be required for any human subjects research. If you obtained IRB approval, you should clearly state this in the paper. 
        \item We recognize that the procedures for this may vary significantly between institutions and locations, and we expect authors to adhere to the NeurIPS Code of Ethics and the guidelines for their institution. 
        \item For initial submissions, do not include any information that would break anonymity (if applicable), such as the institution conducting the review.
    \end{itemize}

\item {\bf Declaration of LLM usage}
    \item[] Question: Does the paper describe the usage of LLMs if it is an important, original, or non-standard component of the core methods in this research? Note that if the LLM is used only for writing, editing, or formatting purposes and does not impact the core methodology, scientific rigorousness, or originality of the research, declaration is not required.
    \item[] Answer: \answerYes{} 
    \item[] Justification: Our method used MLLMs as baseline for image reasoning and segmentation.
    \item[] Guidelines:
    \begin{itemize}
        \item The answer NA means that the core method development in this research does not involve LLMs as any important, original, or non-standard components.
        \item Please refer to our LLM policy (\url{https://neurips.cc/Conferences/2025/LLM}) for what should or should not be described.
    \end{itemize}

\end{enumerate}
\clearpage

\appendix
\renewcommand\thefigure{\Alph{section}\arabic{figure}} 
\renewcommand\thetable{\Alph{section}\arabic{table}} 

\section{Appendix}

In this section, we provide additional figures and tables of the analysis on the proposed method and benchmark. 

\subsection{More Analysis on \textsc{FineRS}-4k}
In this paper, we introduce a new dataset, \textsc{FineRS}-4k, which consists of high-resolution images containing ultra-small objects with diverse spatial distributions. Fig.~\ref{Ap:spatial_distribution} illustrates the detailed distribution of object sizes and spatial locations across all samples in the training, validation, and test sets. As shown, our dataset exhibits more challenging scenarios with extra-small objects and sparse distributions.  
\setcounter{figure}{0}  
\begin{figure}[h]
  \centering
  \includegraphics[width=1\linewidth]{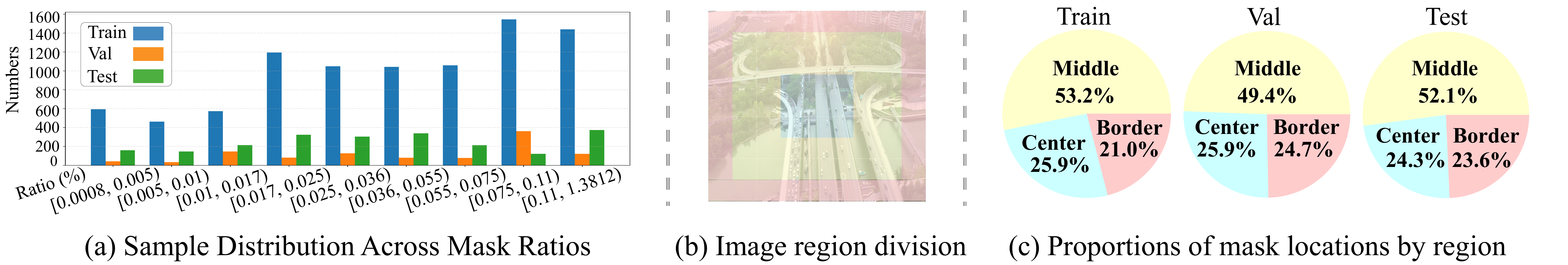}
  \caption{(a) Distribution of mask sizes across three subsets, where the x-axis indicates the ratio of the mask area to the entire image. (b) Spatial division of the image into three regions --- center, middle, and border—based on the location of the mask. (c) Distribution of mask locations across three subsets.}
  \label{Ap:spatial_distribution}
\end{figure}

\subsection{Rewards Definition Details}
In Eq.~\ref{eq:RLPR} and Eq.~\ref{eq:RGSE}, we introduce several rewards for LPR and GSE modules. Specifically, the detailed definitions of each rewards in Eq.~\ref{eq:RLPR} are:
\begin{itemize}
    \item $R_{point}=1$ if L1 distance between predicted point and GT point is less than 100 pixels. 
    \item $R_{b^{L1}}=1$ if the L1 distance between the predicted box and GT box is less than 10 pixels.
    \item $R_{b^{IoU}}=1$ if their IoU is greater than 0.5.
    \item $R_{response}$ is set to 1 if the predicted answer is correct (exact match for multiple choice, fuzzy match for open-ended QA).
    \item $R_{format}$ and $R_{think}$ are binary rewards that verify whether the output adheres to the expected JSON and reasoning formats.
\end{itemize}

In Eq.~\ref{eq:RGSE}:
\begin{itemize}
    \item $R_{{region}^{L1}}=1$ if the L1 distance between the predicted coarse box $B_r^{pre}$ and GT region $B_r^{gt}$ is less than 10 pixels.
    \item $R_{region^{IoU}}=1$ if their IoU is greater than 0.5; $R_{size}$ is 1 when the predicted coarse box is of size $512 \times 512$.
    \item $R_{cover}$ is 1 when the ground-truth object lies fully inside the predicted region.
\end{itemize} 

We assign equal weights to all binary terms, following standard GRPO practices~\cite{liu2025segzero,liu2025visualrft}, which yielded stable performance without tuning.

\subsection{More Comparison Results}
Tab.~\ref{Aptab:2} illustrates the comparison results on the validation set of \textsc{FineRS}-4k. While our method performs slightly lower than SegZero on small-sized objects, it significantly outperforms SegZero on xs and xxs objects. More importantly, our method supports VQA tasks and outperforms other approaches in this setting. Additional qualitative results produced by our method are shown in Fig.~\ref{Ap:visual}.  
\setcounter{table}{0}
\begin{table*}[h]
\scriptsize
\centering
\caption{Performance comparison on the validation set of \textsc{FineRS}-4k. ``$\dagger$'' indicates that the corresponding method is retrained with our dataset. We label the best results with a \textbf{bold} style.}
\begin{tabular}{|l||
    >{\centering\arraybackslash}p{0.6cm} 
    >{\centering\arraybackslash}p{0.6cm} 
    >{\centering\arraybackslash}p{0.6cm}
    >{\centering\arraybackslash}p{0.8cm}|
    >{\centering\arraybackslash}p{0.6cm}
    >{\centering\arraybackslash}p{0.6cm} 
    >{\centering\arraybackslash}p{0.6cm}
    >{\centering\arraybackslash}p{0.6cm}
    >{\centering\arraybackslash}p{0.8cm}|
}
\hline
\rowcolor{gray!45} ~  & \multicolumn{4}{c|}{IoU (gIoU/cIoU)} &  \multicolumn{5}{c|}{QA Acc. (Option/Open)} \\ 

\rowcolor{gray!45} ~~~~~~~~~~~~~~~~~~~~~ \multirow{-2}{*}{Method}  & \textit{S} & \textit{xS} & \textit{xxS} & \textit{All} & \textit{Color} & \textit{Shape} & \textit{Others} & \textit{Position} & \textit{All} \\
\hline
\hline
\rowcolor{gray!10}
\multicolumn{10}{|l|}{\textit{Training-free}} \\

LISA \tiny\texttt{7B}~\cite{lai2024lisa} &14.3/2.91 &6.40/1.06&3.54/0.36&6.58/1.65&0.00/15.1&
0.00/0.00&
0.00/0.00&
0.00/16.7&
0.00/11.6 \\
LISA \tiny\texttt{13B}~\cite{lai2024lisa} &12.1/2.42 &4.10/0.58 &1.21/0.14 &4.28/1.10& 0.00/19.4&
0.00/2.22&
0.00/0.00&
0.00/16.7&
0.00/15.2\\
LISA++ \tiny\texttt{7B}~\cite{yang2023lisa++} &25.0/7.10&8.80/2.10&2.30/0.60&8.91/3.72& 4.86/16.7&
4.86/0.00&
4.00/7.80&
8.33/16.7&
5.99/13.2\\

PixelLM \tiny\texttt{7B}~\cite{ren2024pixellm} &10.5/3.31 &3.30/1.00&0.42/0.13&3.31/1.60 & 0.00/3.78&
0.00/2.22&
0.00/0.00&
0.00/0.00&
0.00/3.21\\
SEAL~\cite{wu2024v} &-- & -- &--  & -- & 
2.16/9.68&
6.98/0.00&
0.00/0.00&
8.33/33.3&
3.20/7.80  \\
DC$^2$ ~\cite{wang2025divide} &-- & -- &--  & -- & 34.6/21.0&
25.6/6.67&
40.0/0.00&
33.3/16.7&
33.2/17.2 \\ 
MLLMs-Know \tiny\texttt{7B}~\cite{zhang2025mllms}  &-- & -- &--  & -- &46.5/50.0&44.2/37.8&80.0/38.5&41.7/50.0&47.2/47.2 \\
MLLMs-Know \tiny\texttt{13B}~\cite{zhang2025mllms} &-- & -- &--  & -- &50.3/51.1&32.6/26.7&80.0/23.1&91.766.7&50.4/45.6\\
MLLMs-Know \tiny\texttt{7B}~\cite{zhang2025mllms} + \scriptsize LISA \tiny \texttt{7B}~\cite{lai2024lisa} &17.1/11.2&8.94/4.64&8.98/2.78&10.4/7.07&46.5/50.0&44.2/37.8&80.0/38.5&41.7/50.0&47.2/47.2 \\
MLLMs-Know \tiny\texttt{13B}~\cite{zhang2025mllms} +\scriptsize LISA \tiny \texttt{13B}~\cite{lai2024lisa}&23.2/16.0&13.5/6.01&9.57/2.55&13.6/9.16&50.3/51.1&32.6/26.7&80.0/23.1&91.7/66.7&50.4/45.6 \\
Seg-zero \tiny\texttt{7B}~\cite{liu2025segzero} &56.5/24.6&28.0/3.75&13.8/1.41&27.0/7.49 & -- & -- & -- & -- & --\\
\hline
        \rowcolor{gray!10}
\multicolumn{10}{|l|}{\textit{Training}} \\
LISA$^\dagger$ \tiny\texttt{7B}~\cite{lai2024lisa} & 14.0/10.8&9.92/7.70&7.27/4.52&9.50/8.62 &4.86/16.7&
2.32/0.00&
39.99/7.69&
8.33/16.7&
5.99/13.2  \\
PixelLM$^\dagger$ \tiny\texttt{7B}~\cite{ren2024pixellm} &1.27/1.02&
0.52/0.35&
0.08/0.02&
0.16/0.13&
0.0/0.0&
0.0/0.0&
0.0/0.0&
0.0/0.0 &
0.0/0.0  \\
MLLMs-Know \tiny\texttt{7B}~\cite{zhang2025mllms} + \scriptsize LISA$^\dagger$ \tiny \texttt{7B}~\cite{lai2024lisa} &1.10/0.72&1.44/1.40&1.77/1.84&1.52/1.05&46.5/50.0&44.2/37.8&80.0/38.5&41.7/50.0&47.2/47.2 \\

Seg-zero$^\dagger$ \tiny\texttt{7B}~\cite{liu2025segzero} &\textbf{67.2}/\textbf{63.8}&45.8/15.8&30.2/16.9&42.9/31.3& -- & -- & -- & -- & --\\
\hline
\hline
\textbf{Ours (\textsc{FineRS})} \tiny \texttt{7B} & {64.8}/{55.3} & \textbf{50.8/27.1} & \textbf{42.5/21.0} & \textbf{49.7/38.6} & \textbf{85.4/65.6} & \textbf{88.4/46.7} & \textbf{100/30.8}& \textbf{91.7/66.7} & \textbf{86.8/60.4}\\
\hline
\end{tabular}

\label{Aptab:2}
\end{table*}

\newpage

\setcounter{figure}{1}  
\begin{figure}[t]
  \centering
  \includegraphics[width=1\linewidth]{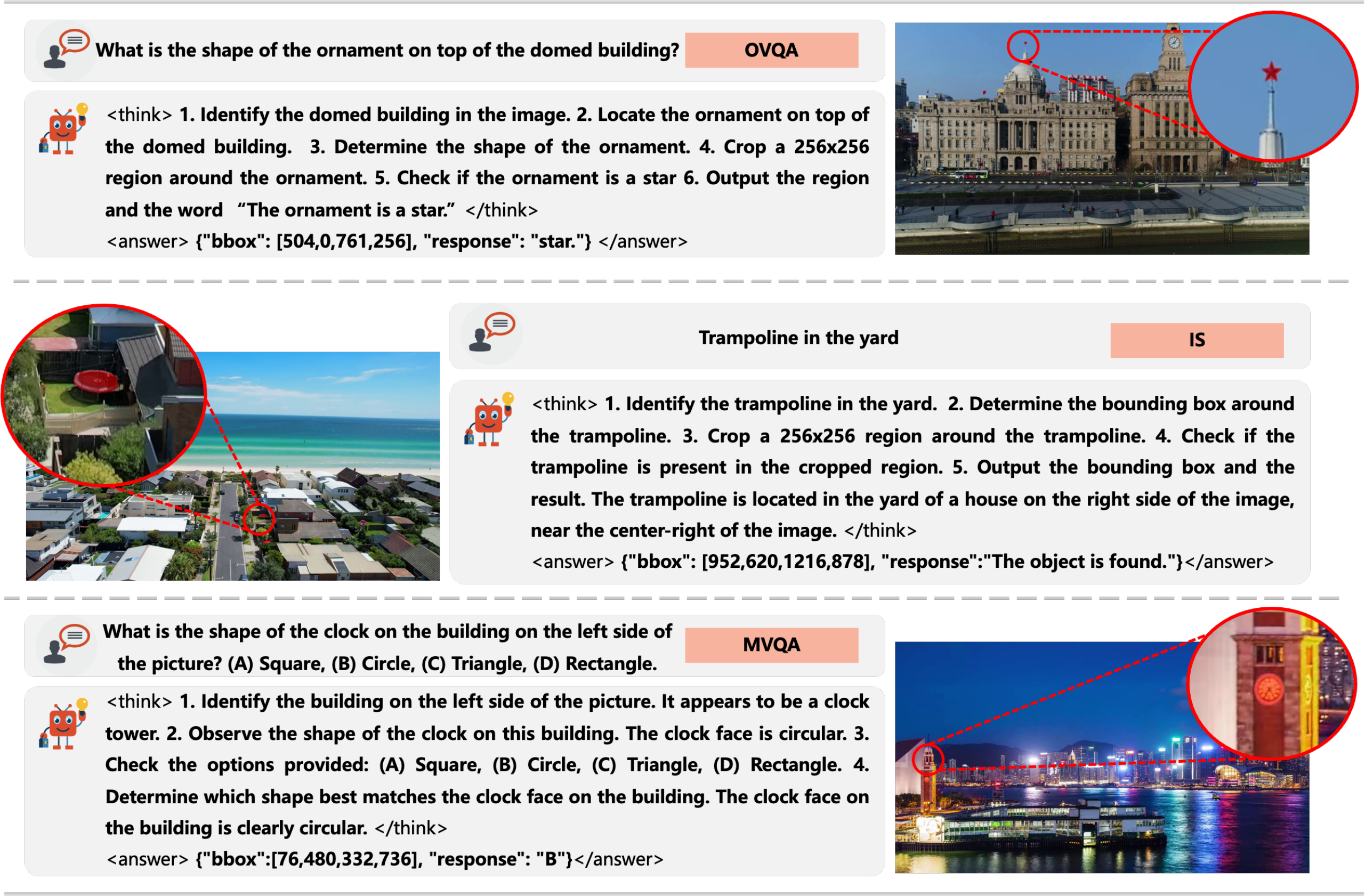}
  \caption{More visual results on Open-ended VQA (OVQA), Multiple-choice VQA (MVQA), and Instruction-guided Segmentation (IS). The listed images are sampled from \textsc{FineRS}-4k test set.}
  
  \label{Ap:visual}
\end{figure}

\subsection{More Ablation Studies}
\textbf{QA Rewards in Different Stages.} Tab.~\ref{ablation2} illustrates the ablation results on the effectiveness of QA accuracy reward and the two-stage designs. To evaluate the efficacy of our two-stage design, we remove the LPR stage and modify the GSE to directly generate object bounding boxes (i.e., $B^{pre}$ in Eq.~\ref{eq:LPR}. This baseline, ``\textbf{One Stage} \textit{xx}'', can be seen as a higher-resolution version of SegZero~\cite{liu2025segzero}. From the comparison results, we can observe that the proposed two-stage coarse-to-fine framework is significantly effective in improving the localization precision of small objects. We also verify the effect of QA accuracy reward in GSE and LPR. The comparison results between ``\textbf{GSE} \textit{w/o QA Acc.} \& \textbf{LPR} \textit{w/o QA Acc.}'' and ``\textsc{FineRS}'' demonstrate that the incorporation of additional VQA ability can effectively enhance the segmentation accuracy on small objects. Besides, the comparison between ``\textbf{GSE} \textit{w/ QA Acc.} \& \textbf{LPR} \textit{w/o QA Acc.}'' and ``\textbf{GSE} \textit{w/o QA Acc.} \& \textbf{LPR} \textit{w/ QA Acc.}'' demonstrates that the textual response on GSE shows better performance due to the exploration of global visual information.

\setcounter{table}{1}
\begin{table*}[h]
\scriptsize
\centering
\caption{Ablation study about QA accuracy reward.}

\begin{tabular}{|l||
    >{\centering\arraybackslash}p{0.6cm} 
    >{\centering\arraybackslash}p{0.7cm}
    >{\centering\arraybackslash}p{0.6cm} 
    >{\centering\arraybackslash}p{0.6cm} |
    >{\centering\arraybackslash}p{0.7cm}
    >{\centering\arraybackslash}p{0.6cm} 
    >{\centering\arraybackslash}p{0.6cm} 
    >{\centering\arraybackslash}p{0.7cm}|
}
\hline
\rowcolor{gray!45} ~ & \multicolumn{4}{c|}{Test set} & \multicolumn{4}{c|}{Val set} \\ 
\rowcolor{gray!45} ~~~~~~~~~~~~~~~~~~~~~~~~\multirow{-2}{*}{Method} ~~~~~~~~ & \textit{gIoU} & \textit{cIoU} & \textit{Option} & \textit{Open} & \textit{gIoU} & \textit{cIoU} & \textit{Option} & \textit{Open}  \\
\hline
\hline

\textbf{One Stage} \textit{w/o QA Acc.} & 46.6 & 38.6& --&-- & 41.6&31.3  &-- & -- \\
\textbf{One Stage} \textit{with QA Acc.}& 42.3 & 41.8 & 84.4 & 58.2& 40.1& 37.6 & 88.5 & 63.6    \\
\textbf{GSE} \textit{w/o QA Acc.} \& \textbf{LPR} \textit{w/o QA Acc.}& 52.8 & 45.7 &  --& -- & 48.6& 37.7 & -- &  --  \\
\textbf{GSE} \textit{w/ QA Acc.} \& \textbf{LPR} \textit{w/o QA Acc.}& 53.7 & 46.2 & {83.3} & {56.7} & 49.2 & 38.2  &87.2&60.0\\
\textbf{GSE} \textit{w/o QA Acc.} \& \textbf{LPR} \textit{w/ QA Acc. }& 54.4 & 46.0 &  56.2& 39.1& 48.1 & 38.2 & 55.6 &  40.4  \\
\textbf{\textsc{FineRS}}  &   \textbf{55.1} & \textbf{46.5}& {83.3} & {56.7}&\textbf{49.9}&  \textbf{39.4} &87.2&60.0  \\
\hline
\end{tabular}

\label{ablation2}
\end{table*}

\begin{wraptable}{r}{0.55\textwidth}
\vspace{-15pt}
\centering
\scriptsize
\caption{Hyper-parameter sensitivity analysis on \textsc{FineRS}-4k test set.}
\begin{tabular}{|l||cccc|}
\hline
  \rowcolor{gray!45}  Hyper-parameters & \textit{gIoU} & \textit{cIoU} & \textit{Option} & \textit{Open} \\
\hline
\hline
    \textbf{Group 8} & \textbf{55.1} & \textbf{46.5} & \textbf{83.3} & \textbf{56.7} \\
    Group 6 & 54.3 & 43.7 & 82.6 & 53.5 \\
    Group 4 & 52.4 & 43.9 & 82.9 & 55.7 \\
\hline
    \textbf{KL 5e-3} & \textbf{55.1} & \textbf{46.5} & \textbf{83.3} & \textbf{56.7} \\
    KL 5e-2 & 54.4 & 46.1 & 82.9 & 55.8 \\
\hline
    \textbf{Seed 42} & \textbf{55.1} & \textbf{46.5} & \textbf{83.3} & \textbf{56.7} \\
    Seed 48 & 53.9 & 45.4 &	82.0 & 56.2 \\
    Seed 80 & 55.1 & 47.6 & 82.9 & 55.7 \\
\hline
\end{tabular}
\label{ap:hyperparams}

\end{wraptable}
\textbf{Effects on Hyper-parameters.} In our experiments, all hyper-parameters are set to the default values in \cite{liu2025segzero} without specific tuning. We conduct a sensitivity analysis on key hyper-parameters, including grouping $n$, KL weight, and seeds. Tab.~\ref{ap:hyperparams} illustrates the representative hyper-parameter results on \textsc{FineRS}-4k test set, which are consistent with SegZero. For a fair comparison, we did not perform expensive hyper-parameter tuning.

\textbf{Efficiency Comparison.} We report the average wall-clock inference time for 4k-resolution inputs in Tab.~\ref{ap:time}. All models were evaluated on a single A100 GPU with consistent runtime environments. As shown, compared to SEAL~\cite{wu2024v} and DC2~\cite{wang2025divide}, our method and Seg-Zero~\cite{liu2025segzero} exhibit higher inference latency due to the use of CoT reasoning. Despite our two-stage framework requires extra inference time, this design is essential for achieving precise reasoning and segmentation of ultra-small objects in high-resolution scenes. 

\begin{table*}[h]
    \scriptsize
    \centering
    \caption{Inference latency and performance of different methods.}
    
    \begin{tabular}{|l||cc|cc|}
    \hline
    \rowcolor{gray!45} & \multicolumn{2}{c|}{\textsc{FineRS}-4k test set} & \multicolumn{2}{c|}{HR-bench 4k}\\
    \rowcolor{gray!45}\multirow{-2}{*}{Method}& \textit{gIoU/cIoU/MVQA/OVQA} & \textit{Time (s/img)} & \textit{QA Acc.} & \textit{Time (s/img)}\\
    \hline
    \hline
    SEAL~\tiny\texttt{7B}~\cite{wu2024v} & --/--/7.53/3.49 & 1.21 & 38.1 & 1.15 \\
    DC$^2$~\tiny\texttt{7B}~\cite{wang2025divide} & --/--/39.2/17.8 & 2.69 & 50.0 & 2.90 \\
    SegZero$^\dagger$~\tiny\texttt{7B}~\cite{liu2025segzero} & 46.4/38.6/--/-- & 8.67 & -- & --   \\
    \hline
    \hline
    Ours (\textsc{FineRS})~\tiny\texttt{7B} & 55.1/46.5/83.3/56.7 & 7.31 (GSE) + 5.57 (LPR\&SAM2) & 63.8 & 6.35 (GSE only)\\
    \hline
    \end{tabular}
    \label{ap:time}
\end{table*}

\textbf{Domain Generalization on ReasonSeg~\cite{lai2024lisa}}. We evaluate our model on ReasonSeg in a zero-shot manner to verify its generalization on non-UVA scenarios. ReasonSeg contains ground-level daily scenes with moderate-to-large object sizes, significantly different from aerial, ultra-high-resolution imagery and ultra-small objects focus of \textsc{FineRS}-4k. We categorize test samples by mask-to-image area ratio into Large (
50\%), Middle (10–50\%), and Small (
10\%). Notably, ReasonSeg's ``Small'' objects are still much larger than FineRS-4k’s (<1\%), introducing a challenging domain and scale gap. Despite this, as shown in Tab.~\ref{ap:reasonseg}, our two-stage model achieves better gIoU on Small objects, outperforming finetuned baselines like LISA$^\dagger$ and Seg-Zero$^\dagger$. This confirms the effectiveness of our coarse-to-fine strategy in segmenting small targets, even under mismatched resolution and context. Moreover, we find that our LPR shows strong adaptability across all sizes and outperforms other methods even without domain-specific finetuning, highlighting its robustness.

\begin{table*}[h]
\centering
\scriptsize
\caption{Performance on the ReasonSeg test set (IoU: gIoU / cIoU). Results are grouped by object size. Notely, ``$\dagger$'' denotes the corresponding methods are finetuned with FineRS-4k without pretraining on large-scale referring segmentation datasets. ``Resize'' means we directly resize ReasonSeg image to meet our model’s require ($1920\times1080$). “Padding” means that the low-resolution image are padded to meet our model’s resolution.}
    \begin{tabular}{|l||cccc|}
    \hline
   \rowcolor{gray!45}Method & Large (316 samples)& Middle (388 samples) & Small (72 samples) & ALL\\
    \hline
    \hline
    Ours (\textsc{FineRS})~\tiny{7B~(Resize)} & 25.2 / 4.92 &	41.4 / 15.6 & 42.3 / 7.12 & 35.0 / 7.16\\
    Ours (\textsc{FineRS})~\tiny{7B~(Padding)} & 27.5 / 5.02&43.2 / 16.1&40.0 / 9.42&36.1 / 7.52\\
    LPR only & 59.0 / 52.4&\textbf{54.4} / \textbf{35.3}&\textbf{45.7} / 24.6&56.6 / 42.1\\
    SegZero$^\dagger$~\tiny{7B~\cite{liu2025segzero}}& 49.3 / 41.7&46.1 / 30.9&35.8 / 15.4&47.1 / 38.8\\
    SegZero~\tiny{7B~\cite{liu2025segzero}}& \textbf{65.3} / \textbf{55.2}&53.5 / 31.4&39.0 / 16.0&\textbf{57.5} / \textbf{52.0}\\
    LISA$^\dagger$~\tiny{7B~\cite{lai2024lisa}}&0.44 / 0.34&4.08 / 1.38&8.14 / 1.12&2.98 / 0.55\\
    LISA~\tiny{7B~\cite{lai2024lisa}}&55.3 / 56.7&34.2 / 26.9&20.3 / \textbf{24.8}&48.7 / 48.8\\
    \hline

    \end{tabular}
    \label{ap:reasonseg}
\end{table*}

\textbf{Results on Other Baselines.} We evaluate our two-stage framework based on Qwen-3b~\cite{yang2024qwen2}, and the results are reported in Tab.~\ref{Aptab:3b}. As observed, employing a smaller backbone leads to noticeable performance degradation, particularly on small objects. Nevertheless, our method still surpasses SegZero~\cite{liu2025segzero} under the same backbone configuration, demonstrating its superior adaptability and robustness.

\begin{table*}[h]
\scriptsize
\centering
\caption{Performance comparison on the test set of \textsc{FineRS}-4k using Qwen2.5-VL (3b). ``$\dagger$'' indicates that the corresponding method is retrained with our dataset. }
\begin{tabular}{|l||
    >{\centering\arraybackslash}p{0.6cm} 
    >{\centering\arraybackslash}p{0.6cm} 
    >{\centering\arraybackslash}p{0.6cm}
    >{\centering\arraybackslash}p{0.8cm}|
    >{\centering\arraybackslash}p{0.6cm}
    >{\centering\arraybackslash}p{0.6cm} 
    >{\centering\arraybackslash}p{0.6cm}
    >{\centering\arraybackslash}p{0.6cm}
    >{\centering\arraybackslash}p{0.8cm}|
}
\hline
\rowcolor{gray!45} ~  & \multicolumn{4}{c|}{IoU (gIoU/cIoU)} &  \multicolumn{5}{c|}{QA Acc. (Option/Open)} \\ 

\rowcolor{gray!45} ~~~~~~~~ \multirow{-2}{*}{Method}  & \textit{S} & \textit{xS} & \textit{xxS} & \textit{All} & \textit{Color} & \textit{Shape} & \textit{Others} & \textit{Position} & \textit{All} \\
\hline
\hline
Seg-zero$^\dagger$ \tiny\texttt{3B}~\cite{liu2025segzero} &\textbf{57.7}/{45.3}&47.4/22.8&26.3/0.86&41.6/29.5& -- & -- & -- & -- & --\\
\textbf{Ours (\textsc{FineRS})} \tiny \texttt{3B} & {57.3}/\textbf{{45.4}} & \textbf{53.8/23.3} & \textbf{43.0/9.86} & \textbf{50.4/29.9} & \textbf{65.6/50.5} & \textbf{68.5/41.0} & \textbf{60.7/21.8}& \textbf{52.6/22.2} & \textbf{65.6/47.0}\\
\hline
\end{tabular}

\label{Aptab:3b}
\end{table*}

\subsection{User Prompt for \textsc{FineRS}}
Fig.~\ref{Ap:GSEprompt} and Fig.~\ref{Ap:LPRprompt} illustrate the user prompts of Global Semantic Exploration (GSE) and Localized Proceptual Refinement (LPR) modules across three tasks, including Instruction-guided Segmentation (IS), Open-ended VQA (OVQA), and Multiple-choice VQA (MVQA).
\setcounter{figure}{2}  
\begin{figure}[h]
  \centering
  \includegraphics[width=1\linewidth]{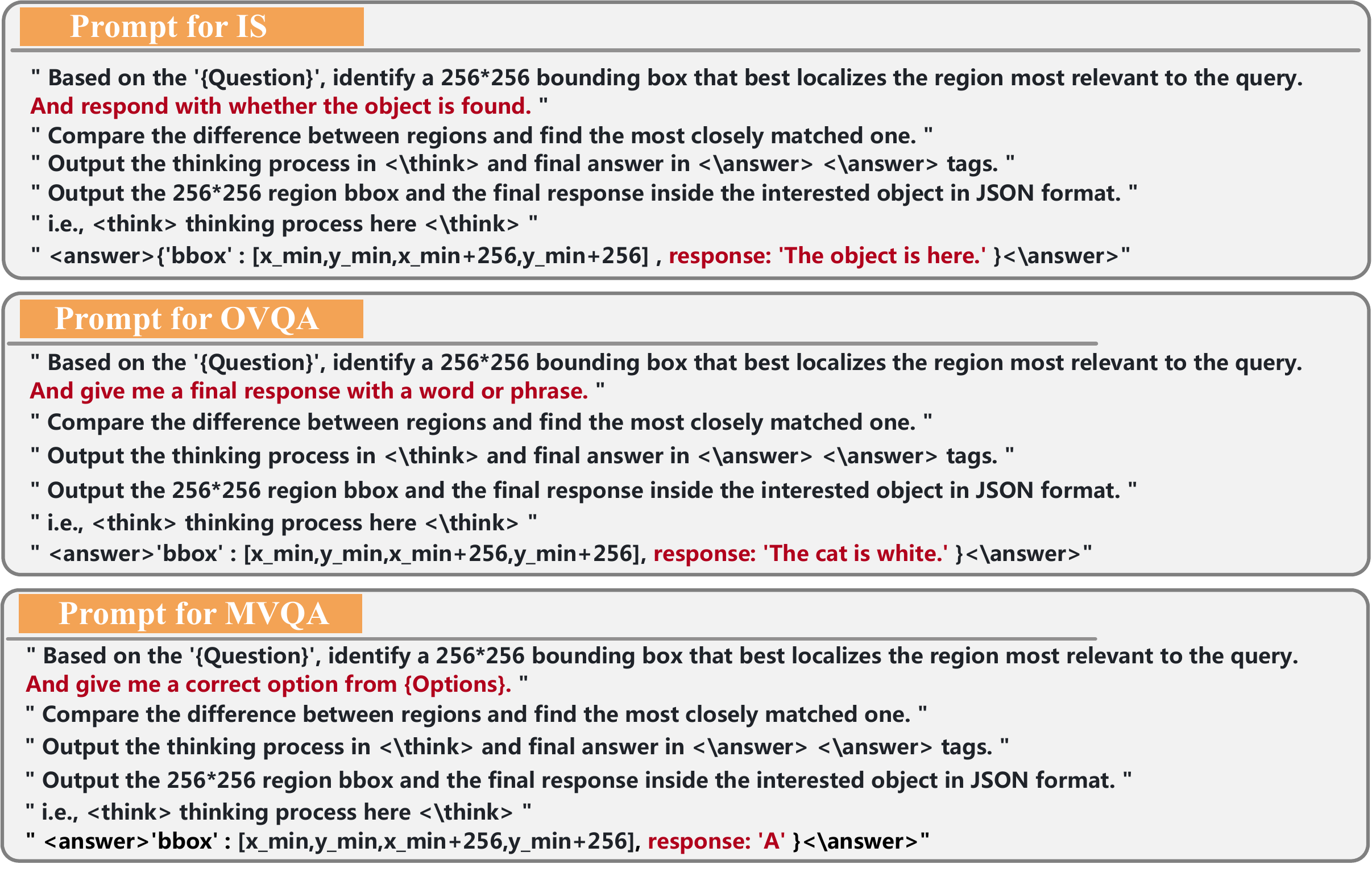}
  \caption{User prompt for GSE module. }
  \label{Ap:GSEprompt}
\end{figure}

\setcounter{figure}{3}  
\begin{figure}[t]
  \centering
  \includegraphics[width=1\linewidth]{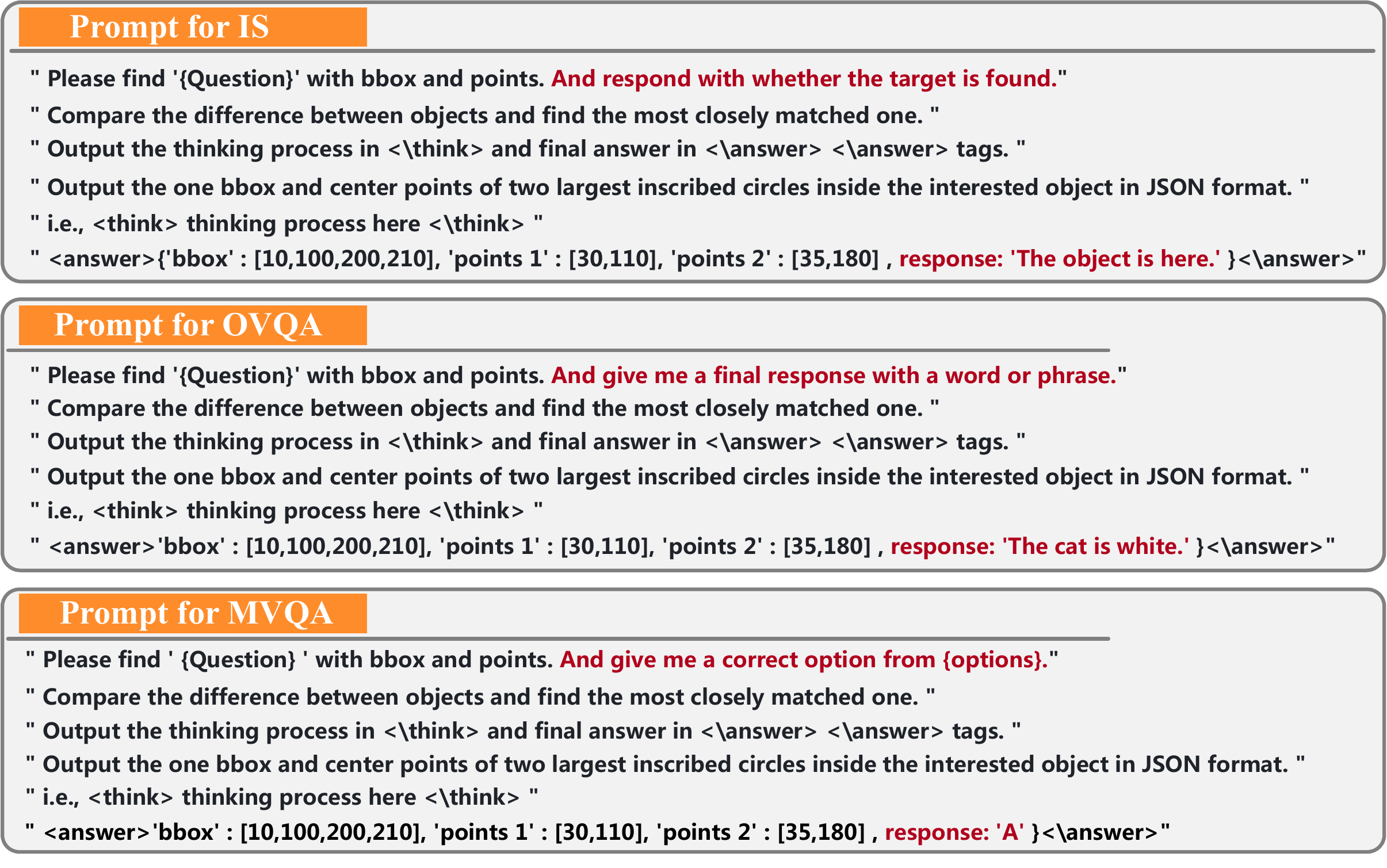}
  \caption{User prompt for LPR module. }
  \label{Ap:LPRprompt}
\end{figure}


\end{document}